\documentclass{article}

\usepackage{PRIMEarxiv}

\usepackage[utf8]{inputenc} 
\usepackage[T1]{fontenc}    

\usepackage{url}            
\usepackage{booktabs}       
\usepackage{amsfonts}       
\usepackage{nicefrac}       
\usepackage{microtype}      
\usepackage{lipsum}
\usepackage{fancyhdr}       
\usepackage{graphicx}       
\usepackage{subcaption}
\usepackage{amsmath}
\usepackage{amssymb}
\usepackage{xcolor}
\usepackage{amsthm}
\graphicspath{{media/}}     
\usepackage{enumitem}

\usepackage{xcolor} 
\usepackage{hyperref}
\definecolor{darkred}{RGB}{150,0,0}
\definecolor{darkgreen}{RGB}{0,150,0}
\definecolor{darkblue}{RGB}{0,0,200}
\hypersetup{
    colorlinks=true, 
    linkcolor=darkred,  
    citecolor=darkblue,   
    urlcolor=cyan   
}

\newtheorem{theorem}{Theorem}
\newtheorem{definition}{Definition}
\newtheorem{lemma}{Lemma}
\newtheorem{proposition}{Proposition}
\newtheorem{fact}{Fact}

\newcommand{\diag}[1]{\operatorname{diag}\left(#1\right)}
\newcommand{\dt}[1]{\frac{\mathrm{d}{#1}}{\mathrm{d}t}}

\newcommand{\eps}{\epsilon}

\newcommand{\R}{\mathbb{R}}

\title{Diagonalizing the Softmax: Hadamard Initialization  for Tractable Cross-Entropy Dynamics
}

\author{
  Connall Garrod \\
  Mathematical Institute \\
  University of Oxford \\
  \texttt{connall.garrod@maths.ox.ac.uk} \\
  \And
  Jonathan P. Keating \\
  Mathematical Institute \\
  University of Oxford \\
  \And
  Christos Thrampoulidis \\
  Department of Electrical and Computer Engineering \\
  University of British Columbia \\
  \texttt{cthrampo@ece.ubc.ca}
}

\begin{document}
\maketitle

\begin{abstract}
{Cross-entropy (CE) training loss dominates deep learning practice, yet existing theory often relies on simplifications—either replacing it with squared loss or restricting to  convex models—that miss essential behavior. CE and squared loss generate fundamentally different dynamics, and convex linear models cannot capture the complexities of non-convex optimization.
We provide an in-depth characterization of multi-class CE optimization dynamics beyond the convex
regime by analyzing a canonical two-layer linear neural network with
standard-basis vectors as inputs—the simplest 
non-convex extension for which the implicit bias remained unknown. This model coincides with the unconstrained features model used to study neural collapse, making our work the first to prove that gradient flow on CE converges to the neural collapse geometry. We construct an explicit Lyapunov function that establishes global convergence, despite the presence of spurious critical points in the non-convex landscape.
A key insight underlying our analysis is an inconspicuous finding: Hadamard Initialization  diagonalizes the softmax operator, freezing the singular vectors of the weight matrices and reducing the dynamics entirely to their singular values. This technique opens a pathway for analyzing CE training dynamics well beyond our specific setting considered here.}
\end{abstract}

\section{Introduction}

\subsection{Motivation}

Modern overparameterized classification networks exhibit a remarkable geometric structure at the end of training, known as neural collapse (NC) \cite{Papyan2020Prevalence}: class features collapse to their respective means, those means form a simplex equiangular tight frame (ETF), and the classifier aligns with this frame. This phenomenon appears consistently across diverse architectures and datasets, despite the absence of any explicit mechanism in the training procedure that enforces it. This raises a central question: \emph{What implicit mechanism in gradient-based optimization produces this precise geometric configuration among the infinitely many solutions that achieve zero training loss?}

This question lies squarely within the implicit bias framework, which seeks to characterize the specific solutions that gradient descent selects when multiple optima exist. For logistic regression,  Soudry et al.~\cite{Soudry_2018} established that gradient descent converges to the max-margin solution. Subsequent research extended this result in several directions, most notably to deep homogeneous networks \cite{lyu2019gradient, Ji_directional, Gunasekar2018ImplicitBO}. These works prove convergence of gradient descent—or its infinitesimal limit, gradient flow—to KKT points of corresponding non-convex max-margin problems. Yet, in contrast to the convex case, these results are comparatively less explicit: they do not specify \emph{which} KKT point is reached.

For NC, Ji et al.~\cite{ji2022an} and Thrampoulidis et al.~\cite{thrampoulidis2022imbalance} applied this framework to the unconstrained features model (UFM), a theoretical proxy for studying representation geometries in deep networks. They showed that cross-entropy (CE) gradient flow converges to a KKT point of a max-margin problem whose global minimizer corresponds to NC. Critical gaps remain, however: these results guarantee only convergence to \emph{some} KKT point—which may be a saddle rather than NC—and provide no characterization of the training trajectory or the convergence rate. More broadly, existing implicit bias results remain restricted to three narrow regimes: (i) logistic regression, which lacks depth; (ii) binary classification, which omits multi-class interactions; and (iii) KKT analyses, which cannot distinguish between NC and other critical points. This gap is particularly significant for unregularized training, where the global minimum is attained only at infinity: analyses based solely on asymptotic limits may overlook crucial finite-time dynamics, such as trajectories remaining near non-NC saddles indefinitely or exhibiting non-monotonic behavior that initially moves away from NC. Understanding implicit bias in this setting requires analyzing the \emph{entire trajectory}, not just its limit.

A complementary line of work, pioneered by Saxe et al.~\cite{Saxe2013ExactST, Saxe2019} and extended by numerous follow-ups, provides tools for such trajectory-level analysis, but only for mean squared error (MSE) loss. Under special initializations that restrict dynamics to singular values, they derived exact solutions for linear networks, showing that singular values evolve independently along sigmoidal curves. This framework was recently applied to the UFM with MSE loss~\cite{zhao2025geometry}. However, NC is observed in classification networks that use CE loss, which introduces a softmax nonlinearity and couples singular value dynamics in ways that prevent exact solutions. Whether the insights and techniques from MSE can extend to CE—and whether they can rigorously characterize the implicit bias toward NC—has remained an open question.

\subsection{Contributions}

We provide the first rigorous proof of convergence to NC under multi-class CE training in a non-convex setting. Our key technical innovation is showing that Hadamard Initialization diagonalizes the softmax operator in the UFM. Concretely, for a normalized Sylvester-Hadamard matrix $U$ and any vector $a$, we prove that:
\[
P(U \diag{a} U^T) = U\diag{ U P(U a)}U^T\,,
\]
where $P(\cdot)$ is the column-wise softmax map. This property enables us to extend the Saxe et al. framework from MSE to CE, reducing its gradient flow to a vector ODE over the $K-1$ non-trivial singular values $a$ of the logit matrix: 
\[
\frac{d a}{dt} = a \odot \big(1- \tilde{U}^T P(\tilde{U}a)\big)\,, 
\] 
where $\odot$ is the entrywise product and $\tilde{U}$ represents the Hadamard modes excluding the trivial constant mode. Unlike the MSE setting, this reduced ODE retains the complex couplings of the softmax: singular values do not evolve independently. Nevertheless, we show the dynamics are sufficiently structured to allow rigorous implicit bias analysis.

\textbf{Non-monotonic convergence.} Prior work has numerically observed monotonic convergence to NC under CE, but we show that this is \emph{not} true in general: convergence can be non-monotonic, and the behavior depends sensitively on the number of classes $K$. For $K \in \{2,4\}$, natural distance metrics decrease monotonically, whereas for $K \geq 8$ we construct explicit initializations where these metrics initially \emph{increase} before eventually decreasing. Even more strikingly, we construct metrics that are monotonic even for $K=8$ but break down for $K \geq 16$. These phenomena are invisible to KKT-based analyses and reveal a subtle, previously unrecognized $K$-dependence in multi-class dynamics. This complexity is missed by prior implicit bias work, which has focused almost exclusively on the binary case.

\textbf{Guaranteed convergence to NC via Lyapunov function.} Despite the non-monotonicity of natural metrics and the presence of competing saddle points, we prove that gradient flow \emph{always} converges to NC by constructing an explicit Lyapunov function—the KL divergence—that decreases monotonically throughout training. This provides the first rigorous characterization of implicit bias toward NC beyond binary classification. Our results demonstrate that NC is not merely an empirical observation but the inevitable outcome of gradient-based optimization in this setting.

\textbf{Exponentially slow convergence.} As the logit norm grows, the CE dynamics slow down exponentially, causing trajectories with unfavorable initializations to linger near non-NC saddle points for arbitrarily long periods. This plateau behavior is practically significant yet invisible to purely asymptotic analyses.

\textbf{Role of initialization.} To reconcile our non-monotonic examples with the monotonic behavior observed in practice, we show that standard near-zero random initialization schemes naturally bias the dynamics toward the simplex ETF subspace, steering trajectories away from problematic saddle regions. This explains why non-monotonic behavior, while theoretically possible, is rarely observed.

\textbf{MSE vs CE comparison.} To highlight the impact of softmax nonlinearity, we show that MSE and CE dynamics coincide only under restrictive conditions: either in the linearized regime (small norms) or under direct logit optimization (removing depth). Outside these settings, CE exhibits coupled dynamics, non-monotonicity, and prolonged saddle plateaus that do not arise under MSE.

Our theoretical findings are supported by numerical experiments.

Although our analysis focuses on Hadamard Initialization in the unregularized, balanced UFM, the approach extends naturally to deeper networks, regularization, and imbalanced classes—directions we leave for future work. Overall, our results provide a tractable framework for understanding optimization dynamics in non-convex multi-class CE settings,
and we hope they motivate the development of similar analytical techniques beyond the UFM.

\section{Background} \label{sec:background}

We consider a classification task with $K$ classes and $n$ samples per class. The $i^{\text{th}}$ sample of class $c$ is denoted by $x_{ic} \in \mathbb{R}^{d_0}$, with corresponding one-hot encoded label $y_c \in \mathbb{R}^K$. A deep neural network $f_\theta (x): \mathbb{R}^{d_0} \rightarrow \mathbb{R}^K$ models the relationship between inputs and labels. We decompose the network as $f_\theta(x) = W h_{\bar{\theta}}(x),$ where $W \in \mathbb{R}^{K \times d}$ are the last layer weights and $h_{\bar{\theta}}: \mathbb{R}^{d_0} \rightarrow \mathbb{R}^d$ is a highly expressive feature map representing the remainder of the network. The parameters $\theta=\{ W,\bar{\theta} \}$ are trained by a variant of gradient descent on the CE loss function. 

We denote the image of the data under the map $h_{\bar{\theta}}$ by $h_{\bar{\theta}}(x_{ic})=h_{ic}$, and define the feature matrix $H$ as the collection of all features in class order: $H = [ h_{1,1}, ... , h_{n,1}, h_{1,2},...,h_{n,K} ] \in \mathbb{R}^{d\times Kn}$. Define the class feature means as $\mu_c=\textrm{Av}_{i}\{ h_{ic} \}$ and the feature mean matrix as $\bar{H} = [\mu_1,\dots,\mu_K]$. We also define the simplex ETF matrix as $$S=I_K - \frac{1}{K} 1_K 1_K^T \in \mathbb{R}^{K \times K}.$$ 

NC then refers to the following phenomenon observed in overparameterized networks as training continues.

\begin{definition} [\textbf{Neural Collapse}]\label{def:NC}
    A neural network exhibits NC if the final-layer weights and feature matrix satisfy:

\begin{itemize}[topsep=-0.5pt,itemsep=-0.5pt]
    \item \textbf{NC1}: Feature vectors collapse onto their class means: $h_{ic}=\mu_c$.
    \item \textbf{NC2}: The  class means align with the simplex ETF: $\bar{H}^T \bar{H}  \propto S$.
    \item \textbf{NC3}: The last-layer weights align with the simplex ETF: $WW^T \propto S$.
\end{itemize}
\end{definition}

The original work by Papyan et al. \cite{Papyan2020Prevalence} included a degree of freedom coming from global centering, but in the context of our model this can be dropped from the definition without impacting results. Our analysis focuses on the non-trivial NC2 and NC3 properties, which dictate the geometry of last-layer weights and features.

\paragraph{Unconstrained Features Models.} The UFM is a proxy used to study the NC phenomenon. To formally define it, we approximate the feature map $h_{\bar{\theta}}(x)$ as being capable of mapping the training data to arbitrary points in feature space, treating the feature vectors $h_{ic}$ as freely optimized variables. Using the CE loss function with no explicit regularization, our objective becomes to minimize jointly over $W\in\R^{K\times d}$ and $H\in\R^{d\times Kn}$:
\begin{equation} \label{eq:CE_UFM}
    \mathcal{L}(W,H) = \sum_{c=1}^K \sum_{i=1}^{n} \log \Big( 1+ \sum_{c' \neq c}^K e^{(Wh_{ic})_{c'}-(Wh_{ic})_c} \Big).
\end{equation}
Under gradient flow, the parameters evolve according to 
\begin{equation} \label{eq:grad_flow}
    \frac{dW}{dt} = (Y-P(WH))H^T, \qquad \frac{dH}{dt}=W^T (Y-P(WH)),
\end{equation}
where $Y=I_K \otimes 1_n^T$, with $\otimes$ denoting the Kronecker product, and $P(WH)$ is the softmax operation applied columnwise to the logit matrix $Z=WH$ of the network, defined as
$$\big[P(Z)\big]_{ij}= \frac{\exp(Z_{ij})}{\sum_{i'} \exp(Z_{i'j})}\,.$$
Using the analysis of \cite{lyu2019gradient}, Ji et al. \cite{ji2022an} showed that this gradient flow converges in direction to a KKT point of the following non-convex max-margin problem:
\begin{equation} \label{eq:max_margin}
    (\hat{W},\hat{H}) \in \textrm{argmin}_{W,H} \Big\{ \frac{1}{2} \| W \|_F^2 + \frac{1}{2} \| H \|_F^2 \Big\}, \quad \textrm{subject to } (w_c - w_{c'})^T h_{ic} \geq 1, \forall c \in [K], i \in [n_c], c' \neq c.
\end{equation}
They further proved that the global minima of this max-margin problem are NC solutions; that is, $\hat{W}$ and $\hat{H}$ satisfy the properties of Definition \ref{def:NC}.  Moreover, all other critical points are saddles.  Thrampoulidis et al. \cite{thrampoulidis2022imbalance} proved that the regularization path of minimizing $\mathcal{L}(W,H)$ with vanishing regularization converges to the global minimizers of Eq. \eqref{eq:max_margin}. They further characterized these global minima in the imbalanced classification setting, showing how solutions deviate from NC geometry as a function of the  imbalance ratio and number of classes.

Despite this progress, the question of implicit bias remains open: whether gradient flow (Eq. \eqref{eq:grad_flow}) converges to these global minima is unknown even in the balanced case, despite the apparent canonical nature of the setting. To understand this gap, note that the UFM is equivalent to a two-layer linear neural network with first layer $H$, second layer $W$, and inputs given by the standard basis vectors in $\mathbb{R}^{Kn}$.

\paragraph{Exact Dynamics with MSE loss.} Replacing the CE loss with MSE loss makes the UFM tractable \cite{zhao2025geometry}. Saxe et al. \cite{Saxe2013ExactST, Saxe2019} introduced such techniques for linear networks trained with MSE loss. Here we show how their methodology applies in the UFM setting with MSE loss. Although their technique is slightly more general, instead being applicable to whitened inputs, the UFM is a canonical instance that as we discussed serves as a meaningful proxy to sufficiently expressive deep neural networks.

Replacing CE loss with MSE loss, the gradient flow equations given in Eq. \eqref{eq:grad_flow} become:
\begin{equation} \label{eq:grad_flow_MSE}
    \frac{dW}{dt}= (Y-WH)H^T, \qquad \frac{dH}{dt}=W^T (Y-WH).
\end{equation}
Let the  singular value decomposition of $Y$ be $Y=U \Sigma V^T$, with singular values $s_c$ for $c=1,\dots,K$. Restrict initialization to a form of spectral initialization: 
$$W=U D_W R^T, \qquad H=R D_H V^T,$$
for some orthogonal matrix $R \in \mathbb{R}^{d \times d}$, with $D_W$ and $D_H$ having their only non-zero diagonal entries being $\alpha_1,...,\alpha_K$ and $\beta_1,..., \beta_K$ respectively. Then, at initialization, the label matrix $Y$ and the logit matrix $WH$ are mutually diagonalizable:
\[
Y-WH = U(\Sigma-D_WD_H)V^T\,,
\]
When put in Eqns. \ref{eq:grad_flow_MSE}, and using $U^TU=V^TV=I$, this implies:
\begin{align}\label{eq:singular evolve mse}
    \frac{dW}{dt}= U\,(\Sigma-D_WD_H)D_H\, R^T, \qquad \frac{dH}{dt}= R\,D_W(\Sigma-D_WD_H)\, V^T.
\end{align}

Thus only the singular values of $W$ and $H$ evolve, while the singular vectors remain fixed. Specifically, from Eq. \eqref{eq:singular evolve mse}, the singular values evolve according to
$$\frac{d \alpha_i}{dt} = (s_i - \alpha_i \beta_i)\, \beta_i, \qquad \frac{d \beta_j}{dt}=(s_i - \alpha_i \beta_i)\, \alpha_i\,.$$
Note that these equations decouple into $K$ independent pairs of ordinary differential equations, and can be solved to show that the products of the singular values $\alpha_j \beta_j$ follow sigmoidal trajectories from their initial value to $s_j$. In particular, this implies that the logit singular values $a_i=\alpha_i\beta_i$ converge monotonically to the singular values of the target matrix $Y$ \cite{Gidel2019}. One can also recover an energy functional corresponding to the dynamics.

\section{Hadamard Initialization} \label{sec:hadamard_initialization}

Contrasting the gradient flow equations for CE and MSE loss, given in Eqns. \eqref{eq:grad_flow} and \eqref{eq:grad_flow_MSE}, highlights the main difficulty in extending the exact dynamics argument to the CE setting: the presence of the softmax matrix $P$. In the MSE case, obtaining exact dynamics requires only that $Y$ and $WH$ be simultaneously diagonalizable, which can be enforced through a suitable initialization. In the CE case, however, one must simultaneously diagonalize $Y$, $WH$, and $P$, which demands precise control over how the softmax nonlinearity interacts with the singular value structure of the logit matrix. Because nonlinear functions are generally unnatural operations on matrices, they rarely preserve such structure.

\subsection{Diagonalizing the Softmax}

Nevertheless, we show that this obstacle can be overcome for CE by initializing with Sylvester Hadamard matrices. 

\begin{definition} [\textbf{Sylvester Hadamard Matrix}] \label{def:sylvester}
    The Sylvester Hadamard matrices $\{ \Phi_{2^m}:m \in \mathbb{N} \}$ are defined recursively by
    $$\Phi_1=1, \quad \Phi_{2^m}=\begin{bmatrix}
    \Phi_{2^{m-1}} & \Phi_{2^{m-1}} \\
    \Phi_{2^{m-1}} & -\Phi_{2^{m-1}}
\end{bmatrix} \in \mathbb{R}^{2^m \times 2^m}.$$
\end{definition}

Each $\Phi_{2^m}$ is a $\pm 1$-valued matrix, whose first row and column consist entirely of ones. Moreover, these matrices  are symmetric $\Phi_{2^m}^T=\Phi_{2^m}$ and satisfy the orthogonality relation $\Phi_{2^m}^T\Phi_{2^m}=2^m I_{2^m}$. Sylvester Hadamard matrices form a special subclass of the more general Hadamard matrices; we leave the study of general matrices to future work.

We will begin by illustrating our technique for the UFM of Eq. \eqref{eq:CE_UFM} in the simplified setting where $n=1$, meaning one data point per class, before establishing the result for general $n$. In this case, the label matrix reduces to  $Y=I_K$, so its singular vector matrices $U$ and $V$ may be taken as any orthogonal matrix with $U=V$. 

We show that by selecting $U= V=\frac{1}{\sqrt{K}} \Phi$, where $\Phi$ is a Sylvester Hadamard matrix, mutual diagonalizability can be achieved even in the presence of the softmax matrix.  \newline

\begin{theorem}
\label{thm:mutual_diag}
    Let $K=2^m$ for $m \in \mathbb{N}$. Suppose $U=\frac{1}{\sqrt{K}} \Phi$, where $\Phi$ is the $K \times K$ Sylvester Hadamard matrix. Denote the columns of $U$ by $u_i$, $i=1,\dots,K$. For a matrix of the form $Z=\sum_i a_i u_i u_i^T$, the corresponding softmax matrix is
$$P(Z)=\sum_i \nu_i u_i u_i^T,$$
\textit{where}
$$\nu_i= \frac{\sum_j \Phi_{ij} e^{\frac{1}{K} (\Phi a)_j}}{\sum_j e^{\frac{1}{K}(\Phi a)_j}}.$$ 
\end{theorem}
Throughout, we refer to this initialization as \emph{Hadamard Initialization}. The theorem proves that this initialization is preserved under the nonlinear softmax operation.
In matrix form, 
the softmax of a matrix diagonalizable by a Sylvester Hadamard matrix remains diagonalizable in the same basis:
\begin{align}\label{eq:matrix thm}
P(U \diag{a} U^T) = U\diag{ U P(U a)}U^T\,,
\end{align}
{where we recall that the softmax map $P(\cdot)$ acts columnwise.} 
 This  diagonalization property follows from the interplay of two fundamental properties of the softmax map and of the Sylvester-Hadamard matrices respectively:
\begin{enumerate}[leftmargin=*]
    \item \emph{Permutation equivariance of the softmax}: For any permutation matrix $\Pi$ and any vector $v$, the softmax map satisfies the identity $P(\Pi v)=\Pi P(v)$.
    \item \emph{Group structure of Hadamard rows}: The rows of $\Phi$, denoted as $\phi_0,\ldots,\phi_{K-1}$, form a group under the Hadamard product (i.e., elementwise multiplication), governed by the relation $\phi_i\odot\phi_j=\phi_{i\oplus j}$, where $\oplus$ denotes the bitwise XOR operation. A direct consequence of this is a key matrix identity: right-multiplying $\Phi$ by a diagonal matrix formed from one of its rows is equivalent to permuting the rows of $\Phi$: $\Phi\diag{\phi_i}=\Pi_i\Phi$, where $\Pi_i$ is the permutation matrix that maps row $j$ to row $i\oplus j$.
\end{enumerate}
When these properties come together, it ensures that the application of softmax to $Z$ (i.e., the LHS of Eq. \eqref{eq:matrix thm}) produces a dyadic matrix: a matrix whose entries are of the form $c_{i \oplus j}$. Such matrices are known to be diagonalized by Sylvester-Hadamard matrices \cite{hadamard_book}, which validates the form given in the equation above and proves the theorem. A detailed proof is provided in Section \ref{sec:mutual_diag_proof}.

\subsection{Tractable CE Dynamics}\label{sec:tractable}

As a consequence of Theorem \ref{thm:mutual_diag}, the arguments of Saxe et al., detailed in Sec. \ref{sec:background}, can be adapted to our setting. Specifically, if the parameter matrices are initialized as $W=U D_W R^T$, $H=R D_H U^T$, for any orthogonal matrix $R \in \mathbb{R}^{d \times d}$, with $U=\frac{1}{\sqrt{K}} \Phi$, then $Y$, $WH$, and $P$ are mutually diagonalizable. This ensures that the singular vectors of $W$ and $H$ are fixed, and all time evolution occurs entirely in their singular values. We emphasize that this property is special to our construction: choosing a general orthogonal matrix for $U$ does not yield mutual diagonalizability of the logits and the softmax matrix. 

Although the previous argument was presented for $n=1$, it extends directly to balanced classes of arbitrary size. This is summarized in the following theorem, a proof of which appears in Appendix \ref{sec:sing_bal_evol_proof}. 

\begin{theorem}
\label{thm:sing_val_evol}
    Let $K=2^m$ for $m \in \mathbb{N}$, and let $U=\frac{1}{\sqrt{K}} \Phi$, where $\Phi$ is the $K \times K$ Sylvester Hadamard matrix. Consider the UFM described in Eq. \eqref{eq:CE_UFM}, with balanced classes of size $n$. Initialize the parameter matrices as $W=UD_W R^T$, and $H=RD_H V^T$, where $R \in \mathbb{R}^{d \times d}$ is orthogonal, $V=U \otimes Q$ where $Q$ is a right singular matrix of $1_n^T$, and the only nonzero singular values of $D_W$ and $D_H$ are $\alpha_1,\dots,\alpha_{K}$ and $\beta_1,\dots, \beta_{K}$, respectively. Then the gradient flow equations, detailed in Eq. \eqref{eq:grad_flow}, reduce to
$$\frac{d \alpha_i}{dt} = \frac{\sqrt{n}}{D} b_i \beta_i, \quad \frac{d \beta_i}{dt} = \frac{\sqrt{n}}{D} b_i \alpha_i,$$
where
\begin{equation} \label{eq:b_def}
    b_i = \sum_{j=1}^{K} [1_K 1_K^T-\Phi]_{ij} \ e^{-\frac{1}{K \sqrt{n}} ([1_K 1_K^T-\Phi] a)_j}, \quad D=\sum_{j=1}^{K} e^{- \frac{1}{K\sqrt{n}} ([1_K 1_K^T - \Phi] a)_j }, \quad \textrm{with } a_i=\alpha_i \beta_i.
\end{equation}
\end{theorem}
Recall that the first row and column of $\Phi$ consist of all ones. Since $[1_K 1_K^T-\Phi]_{1j}=0$ for all $j$, it follows that $b_1=0$, and therefore $\alpha_1,\beta_1$ remain constant. Moreover, since $[1_K 1_K^T-\Phi]_{j1}=0$ for all $j$, they do not appear in the evolution equations for the other singular values. Without loss of generality, we set $\alpha_1=\beta_1=0$ and drop these from our vectors. Thus, from now on we view $\alpha,\beta \in \mathbb{R}^{K-1}$ to be vectors containing only the non-trivial singular values, where the indices have been updated to account for the dropping of the trivial singular mode. 

In keeping with this, we define $\Psi \in \mathbb{R}^{K-1 \times K-1}$ by deleting the first column and row of the matrix $1_K 1_K^T - \Phi$, meaning $\Psi = 1_{K-1} 1_{K-1}^T-X$, where $X=\Phi[2:K,2:K]$ is the core of the Sylvester Hadamard matrix. Note that $-\Psi$ is the Schur complement of the full Hadamard matrix $\Phi$.

It is straightforward to verify that these dynamics correspond to gradient flow on the energy functional
\begin{equation} \label{eq:energy_func}
    E=Kn \log \Bigg( 1+ \sum_{i=1}^{K-1} \exp \Big( - \frac{1}{K\sqrt{n}} (\Psi a)_i \Big) \Bigg),
\end{equation}
where $a_i = \alpha_i \beta_i$.  A direct consequence of this variational structure is the continuous symmetry $\alpha_i \rightarrow \gamma \alpha_i$, $\beta_i \rightarrow \gamma^{-1} \beta_i$, from which Noether's theorem yields the conserved quantities $\alpha_i^2 - \beta_i^2 = C_i$. 

For simplicity, and following common practice in the literature, we set $C_i=0$, which corresponds to the case $\alpha_i = \beta_i$. This reduction is motivated by the fact that small initialization naturally leads to $C_i \approx 0$ in practice. Under this mild assumption, the dynamics can be reduced to the evolution of the \emph{logit singular values} $a_i = \alpha_i \beta_i$, and our differential equations become the following for $i=1,...,K-1$:
$$\frac{da_i}{dt}= \frac{d \alpha_i}{dt} \beta_i + \alpha_i \frac{d \beta_i}{dt} = \frac{\sqrt{n}}{D} b_i (\alpha_i^2 + \beta_i^2) = \frac{2 \sqrt{n}}{D}b_ia_i,$$
We will also absorb the $2 \sqrt{n}$ factor into the time variable, and absorb a factor of $1/(K \sqrt{n})$ into the $a$ variable, which reduces us to the following dynamics for $a \in \mathbb{R}^{K-1}$:
\begin{equation} \label{eq:logit_sing_evol}
    \frac{da_i}{dt} = \frac{1}{D} b_i a_i\,,
\end{equation}
where
\begin{equation} \label{eq:b_and_D_def}
    b_i = \sum_{j=1}^{K-1} \Psi_{ij} e^{-(\Psi a)_j}, \quad D=1 + \sum_{j=1}^{K-1} e^{- (\Psi a)_j }.
\end{equation}

\subsection{Properties of the Reduced Dynamics}
We provide a detailed analysis of the dynamics in Eq. \eqref{eq:logit_sing_evol} in Sec. \ref{sec:implicit bias}. Here, we first establish several basic but important properties of these dynamics that will be used in the subsequent analysis.

\paragraph{On $K=2^m$.} Our construction requires $K = 2^m$ for some $m \in \mathbb{N}$ to exploit the Sylvester-Hadamard structure. Beyond the special case of $K=2$ (binary classification), the dynamics of the CE UFM have not previously been amenable to analysis. Our construction enables tractable analysis for multiple values of $K$, capturing genuine multiclass effects. We therefore view the restriction to $K=2^m$ not as a limitation, but as a modeling choice that enables fine-grained characterization of the dynamics.

\paragraph{Coupled dynamics.} Unlike in the MSE case, the terms $b_i/D$ depend on the full set of singular values $a_1, \ldots, a_{K-1}$. The equations therefore do not decouple, and each singular value influences the trajectory of the others. This additional coupling introduces a level of sophistication absent from the MSE setting and underlies many of the departures from its behavior. 
Because of this lack of decoupling, there are not enough conserved quantities to guarantee closed-form dynamics. Nevertheless, the reduced dynamics in Eq. \eqref{eq:logit_sing_evol} are sufficiently tractable to characterize the implicit bias and convergence properties, as we demonstrate in the subsequent sections.

\paragraph{Positive singular values.} The singular values $a_i(t)$ are monotonically increasing in time. Therefore, if $a_i(0)>0$, then $a_i(t)>0$ for all $t \geq 0$. Moreover, if $a_i(0)=0$, then it remains zero throughout training. We henceforth assume all singular values are initialized to positive values.

\paragraph{Mirror-flow interpretation.} Since $a_i(t)$ remains positive for all future times, we can equivalently write the updates (absorbing the $K,n$-dependent factors) as $
\mathrm{d}(\log(a_i))/\mathrm{d}t = -\nabla_{a_i} E(a)\,,
$ where $E=E(a)$ is the energy functional in Eq. \eqref{eq:energy_func}. In this form, the dynamics are viewed as mirror-flow updates with respect to the entropy potential for the energy functional in Eq. \eqref{eq:energy_func}; see Appendix~\ref{sec:further_theory} for details. 

\paragraph{Normalized singular values.} From Eq. \eqref{eq:logit_sing_evol}, it is straightforward to show that the norm of the logit singular values (e.g., the $L_1$-norm $\|a\|_1=\sum_{i}a_i$) diverges to infinity. This divergence implies that there are no finite minima of the UFM objective on the optimization surface. We therefore study the dynamics of the \emph{normalized logit singular values} $\hat a_i=a_i/\|a\|_1$, which evolve as:
\begin{equation} \label{eq:normalized_dynamics}
    \frac{d}{dt} ( \hat{a}_i) = \frac{\hat{a}_i}{D} \left[ b_i - \bar{b} \right], \quad \textrm{where } \bar{b}= \sum_{j=1}^{K-1} \hat{a}_j b_j. 
\end{equation}
The updates in Eq. \eqref{eq:normalized_dynamics} correspond to replicator dynamics in evolutionary game-theory (see, e.g., \cite{cressman2014replicator} and references therein), where $\hat {a}_i$ represents the strategy of player $i$, $b_i$ is that strategy's payoff, and $\bar{b}$ the average payoff.
Note that the normalized logit singular value vector $\hat{a}$ lies on the $K-1$ dimensional simplex, since with positive initialization all entries remain positive throughout the dynamics.

\section{CE Dynamics and Implicit Bias}\label{sec:implicit bias}

\subsection{NC as a Candidate Among Stable Directions}

{Having established simplified dynamics for UFM gradient flow in Sec. \ref{sec:tractable}, we now analyze their behavior. In particular, we study their \emph{implicit bias}: among the many directions that achieve zero loss under the energy functional in Eq. \eqref{eq:energy_func}, which direction do the dynamics converge to?
Note that any direction $\hat{a}$ satisfying $(\Psi \hat{a})_i >0$ for $i\in[K-1]$ can attain arbitrarily small loss by taking $\|a\|_1 \to \infty$. The key question is therefore which of these zero-loss directions gradient flow in Eq. \eqref{eq:normalized_dynamics} selects.}

{To gain insight into the structure of potential convergence directions, we first identify the \emph{stable directions}, i.e., those satisfying $\dt{\hat{a}}=0$. If the limiting direction satisfies $\hat{a}_i > 0$ for all $i$ (meaning all $K-1$ logit singular values are positive), then the solution is full-rank; otherwise, convergence occurs to a low-rank solution. Among full-rank stable directions, the unique solution is $\hat{a}=\frac{1}{K-1}1_{K-1}$. Indeed, combining $\dt{\hat{a}}=0$ with $\hat{a}_i>0$ for all $i$ implies $b=\bar{b} 1_{K-1}\propto 1_{K-1}$, from which a short algebraic argument (see Appendix \ref{sec:Nc only full rank}) yields the claim.}

{Note that $\hat{a} = \frac{1}{K-1} {1}_{K-1}$ corresponds exactly to the NC solution of Definition \ref{def:NC}. Indeed, equal logit singular values (i.e., $a_i = a_j$ for all $i,j\in[K-1]$) imply that the singular values of $W$ and $H$ are equal as well: $\alpha_i=\beta_i=\alpha_j=\beta_j$ for all $i,j\in[K-1]$. Recall that we have assumed, without loss of generality, that the singular value corresponding to the all-ones eigenspace is zero for both matrices. Together, these conditions yield $W\propto \Phi\,\text{diag}(0,1,\ldots,1) R^T$ and $H\propto R\,\text{diag}(0,1,\ldots,1)V^T$, which in turn give $WW^T \propto S$ and $H^TH \propto S$, where $S=\frac{1}{K}\Phi\,\text{diag}(0,1,\ldots,1)\Phi^T$ is the simplex ETF matrix. This establishes the NC geometry.}

\subsection{Convergence to Neural Collapse via a  Lyapunov Function}
Does gradient flow exhibit an implicit preference for the full-rank NC solution $\hat{a} = \frac{1}{K-1} 1_{K-1}$ among all stable points?
We show that this is indeed the case. Moreover, we establish a stronger result than asymptotic convergence: we identify a Lyapunov function that decreases monotonically along the gradient flow trajectory and explicitly measures the distance from the NC solution.

\begin{theorem} \label{thm:NC_conv}
    Let $K=2^m$, where $m>0$ is a natural number. Let $a(t) \in \mathbb{R}^{K-1}$ be a variable initialized with $a_i(0)>0$ for all $i$, that evolves under the time evolution of Eq. \eqref{eq:logit_sing_evol}. Then the metric $D_{\mathrm{KL}}(\frac{1}{K-1} 1_{K-1} \,\|\, \hat{a})$, where $D_{\mathrm{KL}}$ is the Kullback-Leibler divergence, serves as a Lyapunov function of the flow. Consequently, the metric is decreasing throughout training and we must have 
    $\hat{a} \to \frac{1}{K-1} 1_{K-1}$ as $t \to \infty$. 
\end{theorem}

The proof appears in Section \ref{sec:NC_conv_proof}. This theorem establishes that gradient flow necessarily converges to NC, providing the first rigorous characterization of implicit bias in the CE UFM beyond binary classification.

Our result is notable in two respects.

First, regarding \emph{implicit bias}, existing results typically establish only asymptotic convergence rather than monotonic behavior via a Lyapunov function. Even when restricted to asymptotic convergence, the only prior result applicable to the CE UFM is by Ji and Telgarsky~\cite{Ji_directional}, which is limited to binary classification ($K=2$). However, the binary case is trivial in our framework: the unique stable direction (of rank $K-1=1$) is NC itself. In contrast, for $K \geq 4$, multiple stable directions exist, making the implicit bias question nontrivial. Indeed, as we show in the next subsection, even $K=4$ versus $K=8$ versus $K>8$ exhibit qualitatively different optimization properties: several candidate Lyapunov functions work for smaller $K$ but admit counterexamples for larger $K$.

Second, regarding \emph{neural collapse}, all previous claims about convergence to NC either used explicit finite regularization \cite{zhu2021geometric}, analyzed the regularization path \cite{thrampoulidis2022imbalance}, or characterized the KKT points of the corresponding non-convex max-margin problem in Eq. \eqref{eq:max_margin} \cite{ji2022an}. To the best of our knowledge, this is the first proof of NC occurring purely as a result of gradient flow implicit bias under CE loss.

 \begin{figure}
     \centering
     \begin{subfigure}{0.33\textwidth}
         \centering
         \includegraphics[width=\textwidth]{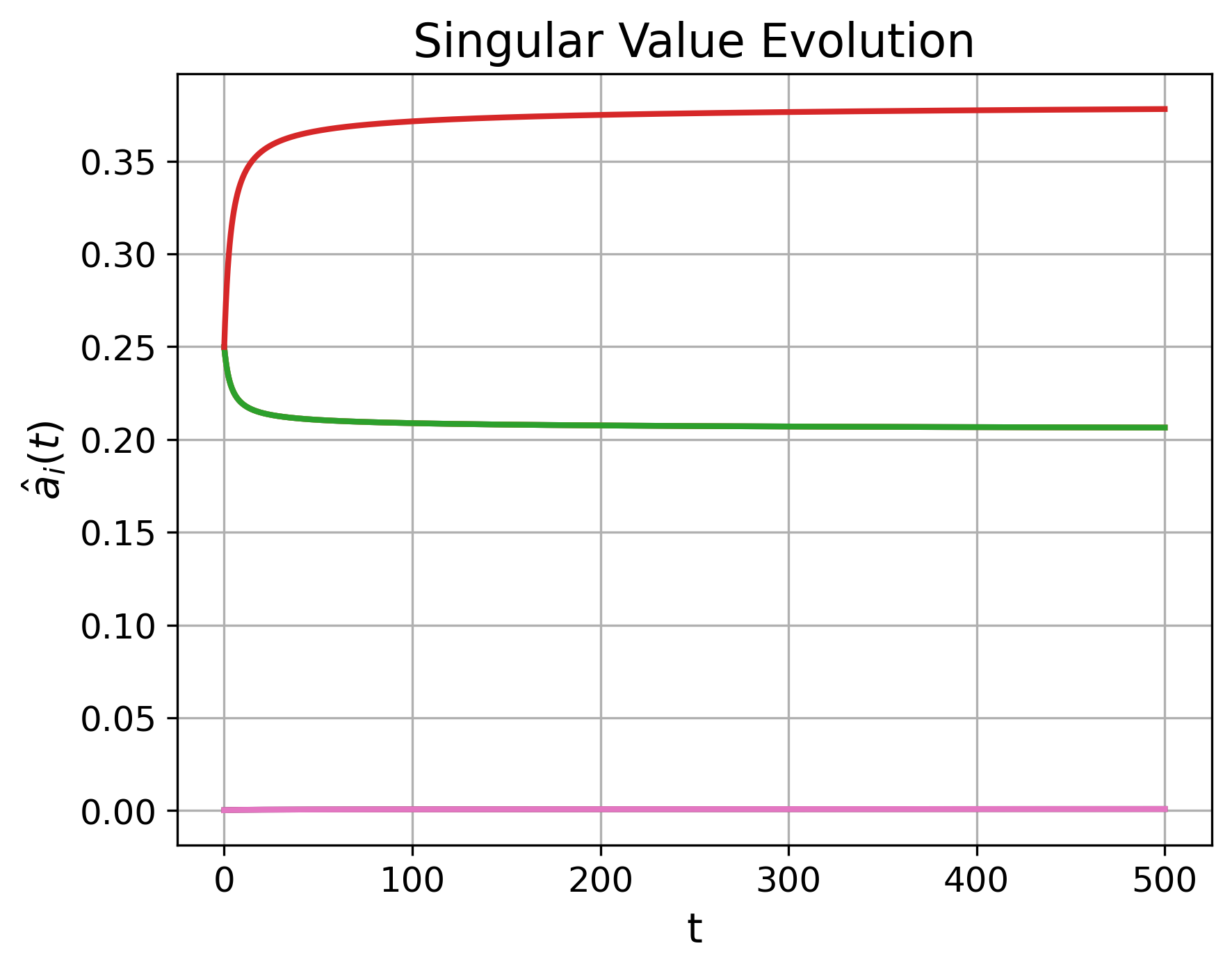}
     \end{subfigure}
     \begin{subfigure}{0.32\textwidth}
         \centering
         \includegraphics[width=\textwidth]{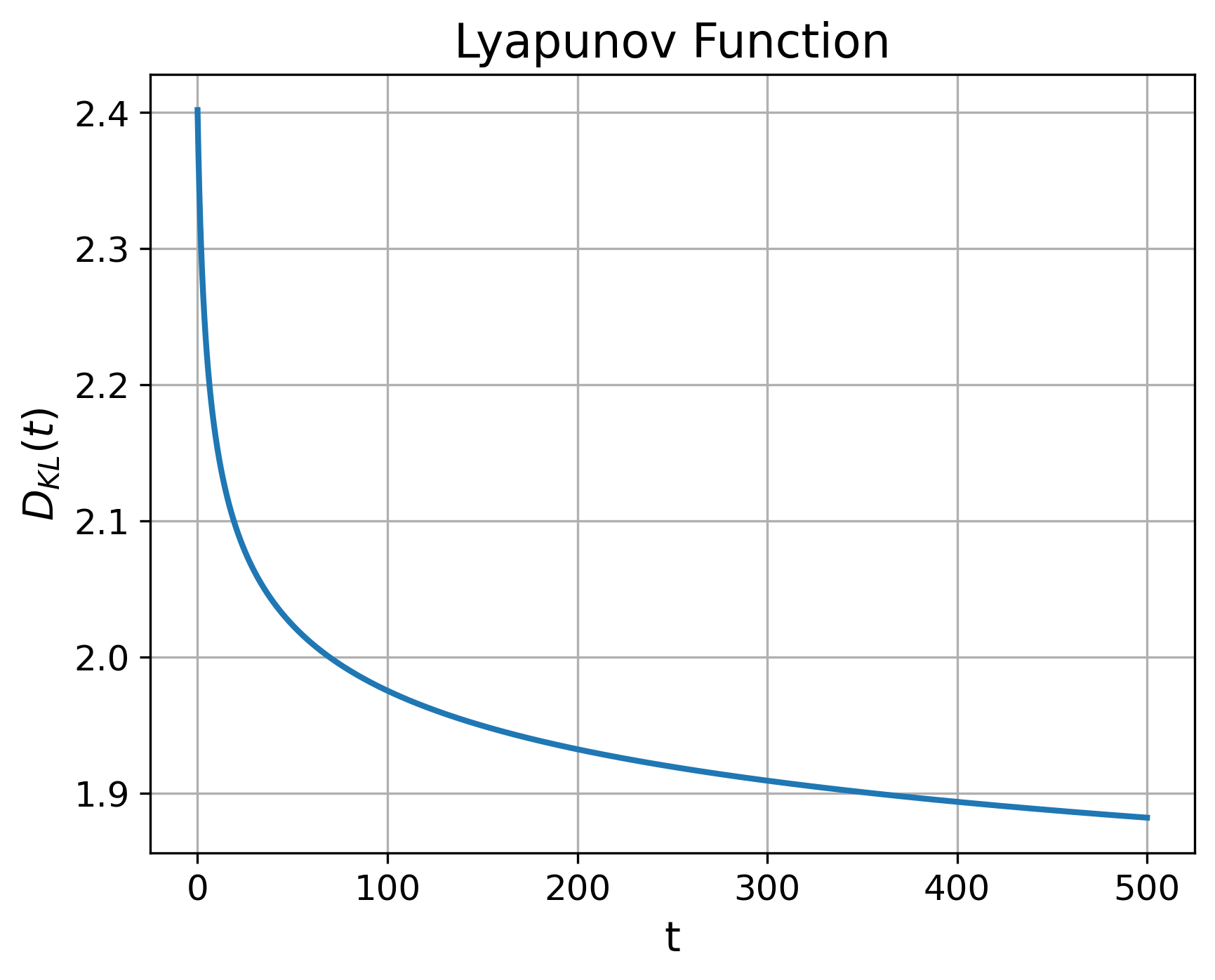}
     \end{subfigure}
     \begin{subfigure}{0.33\textwidth}
         \centering
         \includegraphics[width=\textwidth]{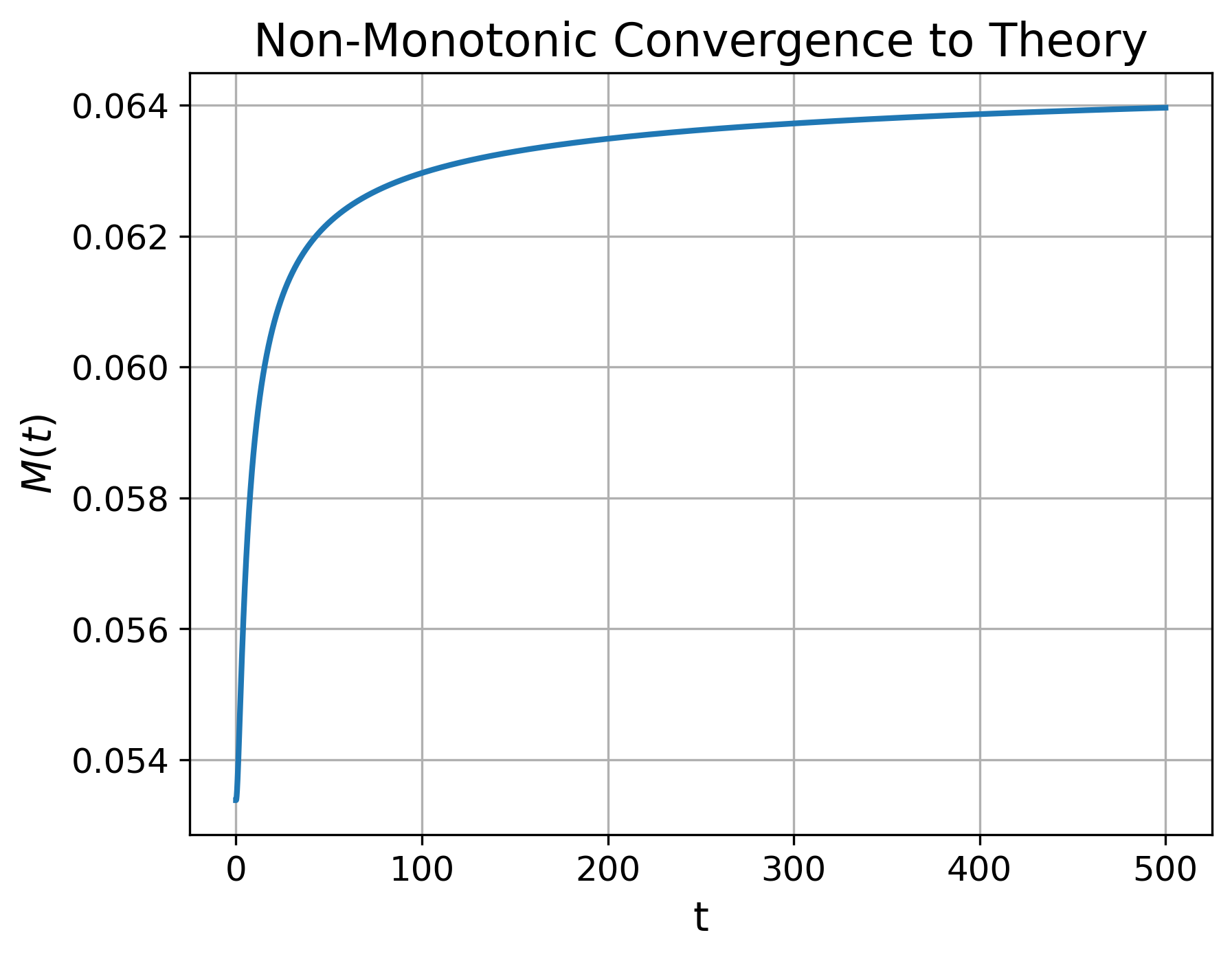}
     \end{subfigure}
     \caption{Training of the logit singular values under the evolution given in Eq. \eqref{eq:logit_sing_evol} with $K=8$, from the initialization $a(0)=[1,1,1,1,0.001,0.001,0.001]$. \textbf{Left:} The evolution of the singular values $\hat{a}_i(t)$. \textbf{Middle:} The evolution of the KL divergence $D_{\mathrm{KL}}(\frac{1}{K-1} 1_{K-1} \,\|\, \hat{a})$. \textbf{Right:} The evolution of the metric $M$, given in Eq. \eqref{eq:metric}. }
     \label{fig:non_monotone_conv}
 \end{figure}

\subsection{Non-monotonic Behaviors}  \label{sec:non_monotone}
The monotonic convergence established in the previous section—via the KL divergence as a Lyapunov function—is a special property. We now show that many other natural distance metrics to NC do not exhibit monotonic behavior under gradient flow.

We begin with the L2-distance metric, which is commonly used in empirical work to quantify distance from NC:
\begin{equation} \label{eq:metric}
    M= \frac{1}{2} \sum_{i=1}^{K-1} \Big( \hat{a}_i - \frac{1}{K-1} \Big)^2\,.
\end{equation}
For $K=2, 4$, this metric decreases monotonically throughout training. However, this monotonicity breaks down for $K \geq 8$, as demonstrated in the following theorem.

\begin{theorem}
\label{thm:non_monotone}
    Let $K=2^m$, where $m>0$ is a natural number. Let $a(t) \in \mathbb{R}^{K-1}$ be a variable initialized with $a_i(0)>0$ for all $i$, that evolves under the time evolution of Eq. \eqref{eq:logit_sing_evol}. If $K=2$ or $K=4$, then the metric $M$, defined in Eq. \eqref{eq:metric}, monotonically converges to 0. If $K \geq 8$, then there exist initializations $a(0)$ such that the metric $M$ is not monotonic. 
\end{theorem}

The proof appears in Appendix \ref{sec:non_monotone_proof}. This result reveals a fundamental qualitative difference from the MSE case studied in~\cite{Saxe2013ExactST, Gidel2019, Saxe2019}: here, the coupled singular values can induce non-monotonic behavior in natural distance metrics.
Figure \ref{fig:non_monotone_conv} illustrates the severity of this phenomenon, showing the evolution of $\hat{a}_i$ and $M$ for an initialization leading to non-monotonic behavior. After a very brief initial decrease, the metric $M$ increases monotonically over the remainder of the considered trajectory, though it must eventually converge to zero. 

\paragraph{The role of non-uniform stable points ($K\geq 8$).} {The mechanism underlying this behavior is as follows: the dynamics first approach a low-rank stable point with non-uniform support—that is, a point where $\hat{a}_i=0$ for some $i\in[K-1]$ and among the non-zero indices, $\hat{a}_i\neq \hat{a}_{j}$ for some $i,j$. Because the trajectory moves toward a non-uniform configuration, metrics such as $M$ measuring uniformity increase.} 
The existence of such non-uniform stable points arises from the structure of the matrix $\Psi$. Our energy functional (Eq. \eqref{eq:energy_func}) depends heavily on the quantities $(\Psi a)_i$, which play the role of margins. Each singular mode $\hat{a}_j$ contributes to enlarging $K/2$ of these margins—precisely those indices $i\in[K-1]$ for which $\Psi_{ij} \neq 0$. As $\|a\|_1$ becomes large, the energy is determined, up to exponentially small terms, by the smallest normalized margins $\arg\min_{i} ( \Psi \hat{a})_i$. While NC maximizes this quantity in the full-rank setting, in rank-constrained regimes we do not necessarily have an equal number of singular values contributing to each margin, making non-symmetric solutions preferable. The case shown in Figure \ref{fig:non_monotone_conv} illustrates this behavior. A more detailed explanation appears in Appendix~\ref{sec:non_uniform_conv_points}.

For $K=4$, all stable points are uniform in their support, preventing such behavior and ensuring a monotonic decrease of $M$. For $K\geq 8$, when non-monotonicity occurs, the optimization path remains near low-rank stable points for extended periods before transitioning to higher-rank saddles, eventually reaching NC. Similar behavior has been observed empirically by Jacot et al.~\cite{jacot2021saddle}, albeit in settings with finite-norm critical points.

\paragraph{Nuanced optimization behavior with respect to $K$.} Beyond this specific example, the $K$-dependence of convergence behavior is more nuanced. The dynamics clearly depend on $K$ through the maximum number of rank transitions along trajectories. Theorem \ref{thm:non_monotone} captures how the nature of convergence changes qualitatively with $K$: from monotonic ($K=4$) to non-monotonic ($K \geq 8$). 
However, this $K$-dependence manifests differently across metrics. The reverse KL divergence $D_{\mathrm{KL}}( \hat{a}  \,\|\, \frac{1}{K-1} 1_{K-1})$ also admits non-monotonic counterexamples for $K=8$ (unlike the KL divergence in Theorem \ref{thm:NC_conv}); see Appendix \ref{sec:examples non-monotonic}. Even more remarkably, some metrics exhibit monotonic convergence for $K=8$ but break down for $K \geq 16$. An example is the logarithm of the ratio between smallest and largest normalized singular values, presented in Appendix \ref{sec:examples non-monotonic}. These findings suggest that many other non-trivial behaviors likely emerge as the class number increases.

\paragraph{Exponential slow down.} The reason behind the potentially excessive timescale of convergence is that lazy training dynamics emerge in the large norm regime. To see this consider the expressions for $b_i/D$ in terms of the variables $\|a\|_1,\hat{a}$:
$$\frac{b_i}{D} = \frac{1}{1+ \sum_{j=1}^{K-1} e^{- \|a\|_1 (\Psi \hat{a})_j }} \sum_{j=1}^{K-1} \Psi_{ij} e^{- \|a\|_1 (\Psi \hat{a})_j }\,.$$
Clearly as $\|a\|_1$ becomes large, these terms decay exponentially in $\|a\|_1$, and so the derivatives of $a_i$ go to zero at $O(\|a\|_1 e^{-C\|a\|_1})$, for some constant $C$, in this limit, and the network evolution becomes incredibly slow. In settings where the loss does not take its minimal value at infinity, such as in the MSE case, these slow down effects do not occur.

\subsection{Further Comparisons to MSE Dynamics}
We have seen that CE dynamics are fundamentally different from MSE dynamics via non-monotonic convergence and slow convergence rates. Here, we further emphasize this difference by identifying the simplifications the CE setting would need to exhibit the simple behavior of MSE.

\subsubsection{Linearized Dynamics} \label{sec:linear_dynamics}
The MSE behavior of independent sigmoidal singular value evolution can be recovered in CE dynamics, but only under a severe restriction: the small $\|a\|_1$ regime where linearized dynamics apply. As the following theorem shows, the singular values decouple in this limit and evolve along sigmoidal trajectories exactly as in the MSE case. However, even with small initialization, the system quickly escape this regime, after which coupling effects introduce the complex behaviors discussed in the preceding subsections. The proof appears in Appendix~\ref{sec:linearized_proof}.

\begin{theorem} \label{thm:linearized}
  Let $K=2^m$, where $m \in \mathbb{N}$. Let $a(t) \in \mathbb{R}^{K-1}$ evolve under the dynamics of Eq. \eqref{eq:logit_sing_evol}, with initialization $a_i(0)>0$ for all $i=1,...,K-1$. To linear order in the scale $\| a\|_1$, the dynamics reduce to
    $$\frac{da_i}{dt}= a_i \left[ 1 - a_i \right] + O \left( \|a\|_1^3 \right).$$
    In particular, the linearized dynamics decouple, and can therefore be solved exactly. The resulting trajectories, that are accurate in the linearized regime, are
    $$a_i(t) = \frac{a_i(0)}{a_i(0)+(1-a_i(0)) e^{-t}}$$
\end{theorem}

The key contrast with MSE is that this decoupled behavior persists at all scales for MSE, whereas for CE it is confined to the small $\|a\|_1$ regime. Moreover, the linearized equations imply $a_i / a_j \to 1$ monotonically, pushing the early-stage dynamics toward NC—though the system exits this regime too quickly for this effect alone to guarantee monotonic convergence to NC.

\subsubsection{Comparison to Logit Optimization}

The second simplification that recovers MSE-like behavior is removing the factorization layer: optimizing logits directly rather than through the product $WH$. This corresponds to the convex one-layer network analyzed by Saxe et al. \cite{Saxe2013ExactST}. The loss in this setting is
\begin{equation} \label{eq:logit_case}
    \mathcal{L}(Z) = \sum_{c=1}^K \sum_{i=1}^{n} \log \Big( 1+ \sum_{c' \neq c}^K e^{(z_{ic})_{c'}-(z_{ic})_c} \Big),
\end{equation}

where $z_{ic}$ are the columns of the logit matrix $Z$ arranged in class order. With Hadamard Initialization  of the logit matrix $Z$, we have the following theorem showing the corresponding reduction in the dynamics, and consequently monotonic convergence to NC.

\begin{theorem} \label{thm:logit_case}
    Let $K=2^m$ for $m \in \mathbb{N}$, and let $U=\frac{1}{\sqrt{K}} \Phi$, where $\Phi$ is the $K \times K$ Sylvester Hadamard matrix. Consider the optimization problem described in Eq. \eqref{eq:logit_case}, and initialize the logit matrix as $Z=UD_Z V^T$, with $V=U \otimes Q$ where $Q$ is a right singular matrix of $1_n^T$, and $D_Z$ has its nonzero singular values given by $a_1,...,a_{K}$. Then the gradient flow equation on this objective, after absorbing a factor of $\frac{1}{K \sqrt{n}}$ into the $a$ variable, a factor of $\frac{1}{K}$ into the $t$ variable, and dropping the trivial singular value, reduces to
    $$\frac{da_i}{dt}= \frac{1}{D} b_i, \qquad i=1,...,K-1.$$
    With the quantities $D$ and $b$ as defined in Eq. \eqref{eq:b_and_D_def}. Moreover, the metric  $M$ in Eq. \eqref{eq:metric} 
    decreases monotonically along trajectories of gradient flow.
    
\end{theorem}

This result parallels that of Soudry et al. \cite{Soudry_2018} for logistic regression, since NC corresponds to the max-margin solution of the UFM. We see that, when optimizing logits directly, the system converges monotonically to the NC solution—just as in the MSE case studied by Saxe et al. \cite{Saxe2013ExactST, Saxe2019}. Taken together, this highlight the role of depth: it is precisely the additional layer that disrupts monotonicity and introduces the more intricate behaviors observed in the CE UFM.

\section{Standard Initialization Schemes lead to NC}
\label{sec:random_good}

The evidence presented in Sec. \ref{sec:non_monotone} raises a natural question: why are such peculiar behaviors not observed in typical training of UFMs \cite{ji2022an} or in standard neural networks? To provide an explanation, we first demonstrate that even though we do not generally see monotonic convergence to NC, the metric $M$ must decrease when it has hit a small enough value.

\begin{theorem} \label{thm:NC_basin}
    Let $K=2^m$, where $m>0$ is a natural number. Let $a(t) \in \mathbb{R}^{K-1}$ be a variable initialized with $a_i(0)>0$ for all $i$, that evolves under the time evolution of Eq. \eqref{eq:logit_sing_evol}. Then the metric $M$, defined in Eq. \eqref{eq:metric}, satisfies the following
    $$M \leq \min \left( \frac{1}{8K^2} , \frac{1}{18K^3 \|a\|_1^2} \right) \implies \frac{dM}{dt} \leq 0$$
    
\end{theorem}

The proof appears in Appendix \ref{sec:NC_basin_proof}. This demonstrates that $M$ will decrease for a finite time training trajectory, provided the initialization is sufficiently close to the NC solution.

We can now explore how typical initialization schemes, which feature random i.i.d. entries in the parameter matrices, impact the singular values of the logit matrix. Note this differs from our modeling context involving Hadamard matrices. However, empirically, we find that our modeling approximates random dynamics from small initialization; see Sec. \ref{sec:experiments}. We show below that with Gaussian initialization at small scale, the zeroth-order dynamics already bias the system towards the simplex ETF. 

\begin{theorem} \label{thm:initialization}
    Consider the UFM defined in Eq. \eqref{eq:CE_UFM} and evolve the parameter matrices under the gradient flow of Eq. \eqref{eq:grad_flow}, with initialization $W(0)=\epsilon d^{-\frac{1}{2}}A_1$, $H(0)=\epsilon d^{-\frac{1}{2}} A_0$, where the matrices $A_1,A_0$ have entries drawn i.i.d. from a standard Gaussian distribution, and $d$ is the network width. Defining the quantity $t^* = \frac{1}{2 \sqrt{n}} \log(\frac{1}{\epsilon})$, and taking the limit $\epsilon \to 0$, the leading order term of $Z(t^*)$ is given by
    $$Z(t^*) \sim \frac{1}{2d \sqrt{n}} \epsilon^2 e^{2t^* \sqrt{n}}\left[ S X X^T (S \otimes 1_n^T)\right], $$
    where $X \in \mathbb{R}^{K \times d}$ has entries given in distribution by independent standard Gaussians. Hence additionally taking the $d \to \infty$ limit, we recover asymptotically
    $$Z(t^*) \sim \frac{1}{2 \sqrt{n}} \epsilon^2 e^{2t^* \sqrt{n}} \left( S \otimes 1_n^T \right).$$ 
\end{theorem}

The theorem shows that for small initialization, after some period of training in the zeroth order regime, the logit matrix approximates a simplex ETF, with the quality of the approximation improving as initialization becomes smaller and the network width increases. Thus, zeroth-order dynamics naturally drive the logit matrix toward the simplex ETF structure. 

Since the linearized dynamics further reinforce this geometry, and the NC solution is stable with an attractive basin at all scales, this naturally leads the trajectory to avoid regions of non-monotonicity that occur around the saddles in the surface. This explains why Ji et al. \cite{ji2022an} observe only monotonic behavior: the initialization scheme itself imposes a strong bias toward the simplex ETF, effectively concealing the more intricate dynamics revealed by our analysis.

\section{Numerical Experiments}\label{sec:experiments}

\subsection{Setup}
We simulate the gradient flow of the UFM (Eq. \eqref{eq:CE_UFM})  using gradient descent with a small step size $0.01$. We set the number of classes $K=4$, the hidden dimension $d=K$, and use (for simplicity) $n=1$ sample per class. We compare two initialization schemes: (1) \emph{Random Init.:} We initialize $W$ and $H$ with i.i.d. entries drawn from a zero-mean Gaussian distribution with variance $1/K$, scaled by a small $\epsilon = e^{-6}/\sqrt{K}\approx 10^{-3}$. (2) \emph{Hadamard Init.:} We initialize $W = U D_W R^T$ and $H = RD_H V^T$ as described in Theorem \ref{thm:mutual_diag}, $U=V=\frac{1}{\sqrt{K}}\Phi$. Recall from Sec. \ref{sec:hadamard_initialization} that the singular value corresponding to the all $1$s column of $U$ is not updated by the flow; thus, as in theory, we assume without loss of generality that the respective entry of $D_W,D_H$ is set to zero. To further ensure direct comparison to Random Init., we set the remaining $K-1$ singular values such that $D_W = D_H$ and they are identical to the largest $K-1$ singular values of the logit matrix from its corresponding Random Init. pair. This isolates the \emph{initial} difference between the two schemes to only the singular vectors. 
Alongside these, and with same initialization, we numerically solve the reduced gradient flow equations for the singular values (Eq. \eqref{eq:logit_sing_evol}), which we label \emph{ODE}. 

All results are averaged over 10 runs, with the shaded regions in our plots representing the standard error of the mean.
We track the following metrics as function of iterations:  (a) Cross-Entropy Loss, (b) classically used NC alignment metrics (the logit one corresponds to the metric $M$ in Eq. \eqref{eq:metric}), (c)
KL Divergence metric of Thm. \ref{thm:NC_conv}, (d) singular values, (e) normalized (by L1-norm) singular values, (f) Residual energy outside the signal subspace of Hadamard Init.. See Appendix \ref{sec:details exp} for details. We log all metrics at $100$ geometrically-spaced iterations (for log-scale visualization).

In Appendix \ref{sec:details exp} we present additional experimental ablations on: (a) larger number of classes ($K\geq 4$), (b) non-infinitesimal initialization scale ($\epsilon=1/\sqrt{K}$), and, (c) overparameterized hidden layer dimension ($d/K>1$). The latter support the theoretical findings of Theorem \ref{thm:initialization}:  overparameterization $d\gg K$ accelerates convergence of the dynamics under infinitesimal random initialization to the simplex ETF geometry.

\subsection{Results}
\begin{figure}[t]
    \centering
    \includegraphics[width=0.9\textwidth]{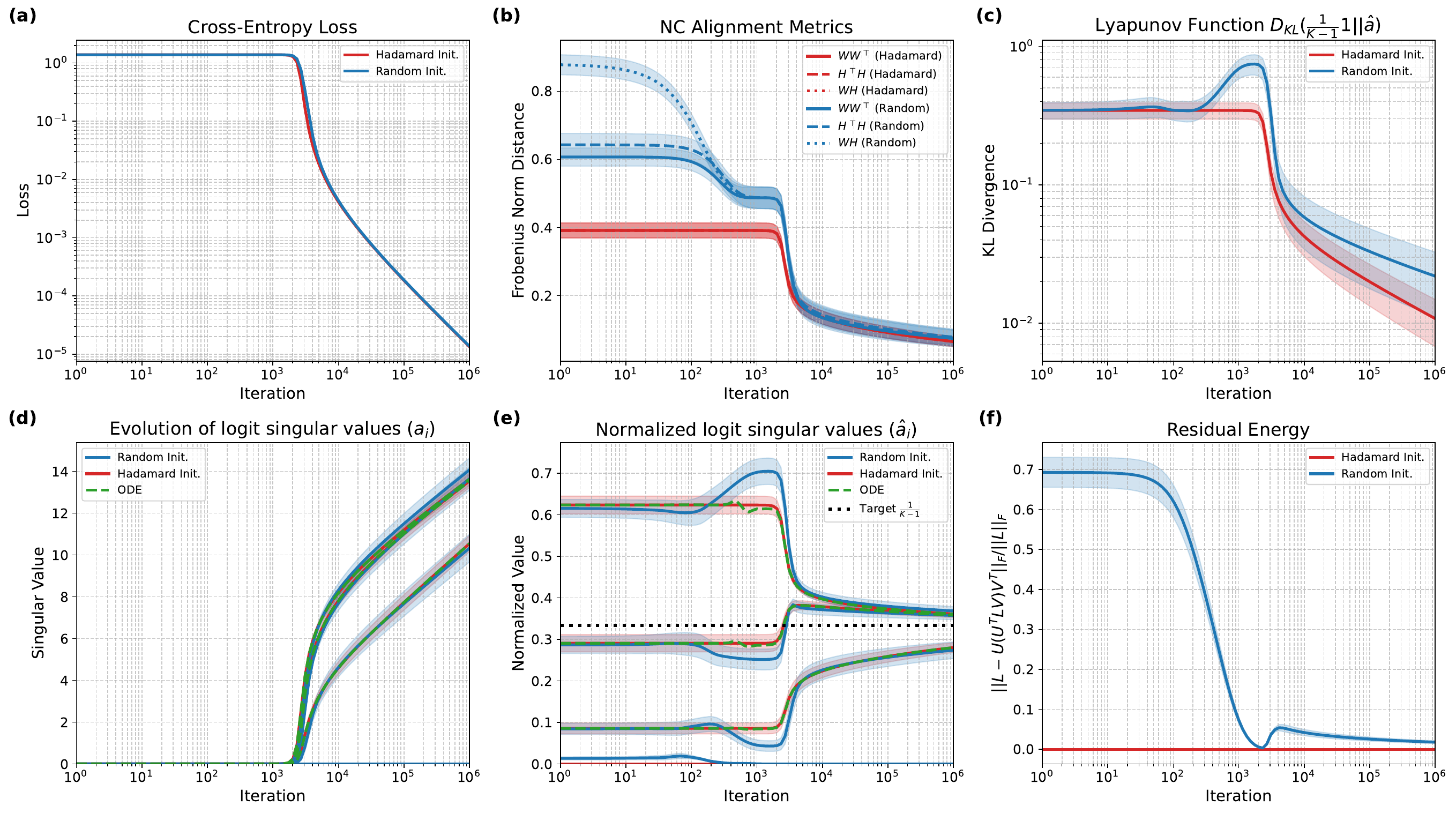}
    \caption{
        \emph{Comparison of Hadamard Init., Random Init., and the theoretical ODE dynamics.} Plots show metrics vs. training iteration on a log scale. Shaded regions denote standard error over 10 realizations. Sec.~\ref{sec:experiments} for detailed discussion.
    }
    \label{fig:V=4_svd_vs_rand}
    \vspace{-0.1in}
\end{figure}

Our results are summarized in Fig.~\ref{fig:V=4_svd_vs_rand}.

\emph{Hadamard Initialization  freezes singular vectors.}
For Hadamard Init., the residual energy in plot (f) remains at its initial zero value for all time. This confirms the central finding of Thm.~\ref{thm:mutual_diag}: Hadamard Initialization  perfectly diagonalizes the softmax, ensuring the singular vectors of $W$ and $H$ remain fixed throughout training.

\emph{Validation of reduced ODE.}
Plots (d) and (e) validate the theoretical reduction in Thm.~\ref{thm:sing_val_evol} of the original matrix ODE (Eq.~\eqref{eq:CE_UFM}) to the simpler vector ODE on the logit singular values (Eq.~\eqref{eq:logit_sing_evol}). The empirical singular values from the Hadamard Init. (red) simulation are perfectly aligned with the ODE solution (green dotted line).

\emph{Convergence to NC.}
From both Hadamard and Random Init., the dynamics successfully converge to the NC geometry. This is captured in Plots (b) and (c) by tracking the standard NC convergence metrics from the literature and our Lyapunov function, respectively. Plot (e) also clearly shows the normalized singular values $\hat{a}_i$ for both initializations converging to the target $\frac{1}{K-1} = \frac{1}{3}$, as predicted by Thm.~\ref{thm:NC_conv}, albeit at a slow rate.

\emph{Both initializations show a qualitatively similar two-stage behavior:}
\begin{enumerate}[leftmargin=*]
\item \emph{Alignment / ``Burn-in'' Stage ($t < 10^3$).} 
In this first stage, both methods show unchanged \emph{Loss} (plot a) and a (near) plateau in the \emph{alignment metrics} (plots b,c). However, their internal dynamics differ:
\begin{itemize}[leftmargin=*]
\item For {Hadamard Init.} (red), which starts perfectly aligned (\emph{Residual Energy} in (f) is zero), the \emph{normalized singular values} (e) and \emph{Lyapunov function} (c) remain perfectly flat.
\item For {Random Init.} (blue), this is an active alignment phase. The \emph{Residual Energy} (f) decays as vectors align to the ETF subspace. This alignment drives a reconfiguration of the \emph{normalized singular values} (e), which are moving but \emph{not} yet converging to their final equilibrium. This reconfiguration, in turn, causes the temporary oscillations seen in the \emph{Lyapunov function} (c).
\end{itemize}
\item \emph{Convergence Stage ($t > 10^3$)}
 The starting point of this stage coincides for both methods. Now, the \emph{normalized singular values} (e) for both methods begin their slow convergence toward their final equilibrium. This ``learning" phase coincides with a sharp increase in singular values magnitudes (d), a sharp decrease in {Loss} (a) and the final decrease in all convergence metrics (plots b and c).
\end{enumerate}

The burn-in stage has a natural interpretation for Random Init.: the dynamics first search for a good subspace; once close to the ETF subspace, the parameter magnitudes grow sharply (plot (d)) to saturate the softmax and reduce the CE loss, while the normalized singular values then drift slowly toward equilibrium. A similar ``waiting'' behavior is visible under Hadamard Init.: even though the subspaces are perfectly aligned from the start, there is initially very slow motion in the singular-value dynamics before the same growth-and-converge pattern sets in. This suggests a long time is needed for the dynamics to ``escape'' the neighborhood of the saddle point $W=H=0$ to which the infinitesimal initialization is close to.

Another interesting dynamic, visible in plots (b) and (c), is the difference in monotonicity between the metrics. For Hadamard Init., both the L2-based metric $M$ (plot b) and the Lyapunov (KL) function (plot c) decrease monotonically, which is consistent with Theorem \ref{thm:NC_conv} and (for $K=4$) Theorem \ref{thm:non_monotone}. For Random Init., however, the story is different: the Lyapunov function (plot c) is non-monotonic, showing oscillations during the alignment phase, even while the L2-based metric $M$ (plot b) is monotonic. As shown in Appendix \ref{sec:details exp}, the non-monotonicity of the Lyapunov function for Random Init. persists for $K=8$ and $K=16$, while the metric $M$ remains (nearly) monotonic for Random Init. even at larger $K$.
This difference is intuitive: outside the spectral setting we analyze, singular vectors impact metric $M$ but not $D_{\textrm{KL}}$. Thus $M$ benefits during the alignment phase from singular vector evolution, while $D_{\textrm{KL}}$ does not. Importantly, this does not contradict Theorem \ref{thm:non_monotone}, which proves the existence of a non-monotonic trajectory. That counterexample requires specific non-uniform initialization, a regime that standard i.i.d. Gaussian initialization naturally avoids (consistent with the conclusions of Sec. \ref{sec:random_good}).

Overall, our findings in the appendix for larger values of $K=8,16$ confirm our conclusions about the presence of two-stage dynamics and qualitative  consistency between random and Hadamard Init..
 However, they also emphasize the intricacies of multi-class CE optimization. We observe that the quantitative approximation of Random Init. by Hadamard Init. deteriorates as $K$ increases (e.g., $K=16$ in Appendix \ref{sec:details exp}). This is yet another departure from the MSE setting \cite{Saxe2013ExactST}, where such spectral approximations are typically very accurate. The softmax non-linearity introduces a complex interaction between singular vector alignment and singular value evolution that invites further investigation.

\section{Proofs}

In this section, we present the proofs of two core theoretical results: that Hadamard initialization diagonalizes the softmax (Theorem~\ref{thm:mutual_diag}) and that the KL divergence characterizes convergence to the neural-collapse geometry (Theorem~\ref{thm:NC_conv}).

\subsection{Proof of Theorem \ref{thm:mutual_diag}} \label{sec:mutual_diag_proof}

Let $\phi_i, i\in\{0,1,\ldots,K-1\}$ be the {rows} of  the $K\times K$ Sylvester Hadamard matrix $\Phi$. The indexing from $0$ to $K-1$ is convenient for presenting the following key circular property of Sylvester Hadamard matrices \cite[Lemma 3.6]{hadamard_book}.
\begin{fact}\label{fact:closed under Hadamard product}
    For any $i,j\in\{0,\ldots,K-1\}$, it holds that 
    $\phi_i\odot\phi_j = \phi_{i\oplus j}\,,$
    where $i\oplus j$ is the bitwise XOR of $i$ and $j$\,.
    \end{fact}

For convenience, define matrices
\[
\Phi_i:=\Phi\diag{\phi_i} = \left[\phi_j^T\odot\phi_i^T\right]_{j\in[k]} \quad\text{ for all } i\in\{0,\ldots,K-1\}.
\]
Note that $\Phi_0=\Phi$. Using Fact \ref{fact:closed under Hadamard product}, we can express $\Phi_i$ in terms of the permutation matrix 
\[
\Pi_i = \left[e^T_{i\oplus j}\right]_{j\in[k]}\,, \quad\text{for all } i\in\{0,1,\ldots,K-1\}
\]
as follows:
\begin{align}\label{eq:Phi i as permutation}
\Phi_i = \Pi_i\Phi,  \quad\text{ for all } i\in\{0,\ldots,K-1\}.
\end{align}
Recall that softmax is applied columnwise. Thus, to prove the advertised identity in Eq. \eqref{eq:matrix thm}, it suffices for any column $i\in\{0,1,\ldots,K-1\}$ that
\[
P(\frac{1}{K}\Phi\diag{a}\phi_i) = \frac{1}{K}\Phi \diag{\Phi P(\frac{1}{K}\Phi a)}\phi_i\,.
\]
Using the property $\diag{u}v=\diag{v}u$ for any $u,v$ on both sides of the equality, and applying Eq. \eqref{eq:Phi i as permutation}, this is equivalent to
\[
P(\frac{1}{K}\Pi_i\Phi a) = \frac{1}{K}\Pi_i\Phi\Phi P(\frac{1}{K}\Phi a)\,.
\]
But, $\Phi\Phi=\Phi\Phi^T=kI_K$, thus the claim reduces to proving
\[
P(\frac{1}{K}\Pi_i\Phi a) = \Pi_i P(\frac{1}{K}\Phi a)\,.
\]
In turn, this identity is true because of the permutation equivariance property of softmax, i.e. $P(\Pi u) = \Pi P(u)$ for any permutation matrix $\Pi$ and vector $u$. This completes the proof.

\subsection{Proof of Theorem \ref{thm:NC_conv}} \label{sec:NC_conv_proof}

The KL divergence $
D_{\mathrm{KL}}(p \,\|\, q) =\sum_i p_i \log \left(\frac{p_i}{q_i} \right)
$ between two categorical distributions with probabilities $p$ and $q$ is  non-negative and equals  zero if and only if $p=q$.
In our setting, we have:
\[
D_{\mathrm{KL}} \left(\frac{1}{K-1} 1_{K-1} \,\|\, \hat{a} \right)=- \log(K-1) - \frac{1}{K-1} \sum_i \log(\hat{a}_i) .
\]
For the remainder of the proof we denote this quantity by $D_{\textrm{KL}}$. Differentiating, we obtain
\begin{align}\label{eq:KL_comp}
\frac{d}{dt} \left( D_{\textrm{KL}} \right) = - \frac{1}{K-1} \sum_i \frac{1}{\hat{a}_i} \frac{d}{dt} (\hat{a}_i)= - \frac{1}{(K-1)D} \left[ \left(\sum_i b_i \right) - (K-1) \bar{b} \right]\,,
\end{align}
where the second equality follows from substituting Eq. \eqref{eq:normalized_dynamics}.

We now define $x_i = (\Psi a)_i$ and rewrite the derivative in terms of these variables, which recall play the role of margins. 
First observe that
\[
\bar{b} = \sum_i \hat{a}_i b_i = \sum_{ij} \hat{a}_i \Psi_{ij} e^{-(\Psi a)_j} = \frac{1}{\|a\|_1} \sum_{ij} a_i \Psi_{ij} e^{-(\Psi a)_j},
\]
which simplifies to
\begin{equation} \label{eq:b_bar_comp}
    \bar{b} = \frac{1}{\|a\|_1} \sum_j x_j e^{-x_j}.
\end{equation}
Second note
\[
\sum_i b_i = \sum_{ij} \Psi_{ij} e^{-(\Psi a)_j}.
\]
Using $\sum_i \Psi_{ij} = K$, we obtain
\[
\sum_i b_i = K \sum_j e^{-(\Psi a)_j},
\]
Moreover, since
\[
\sum_i (\Psi \hat{a})_i = \sum_{ij} \Psi_{ij} \hat{a}_j = K \sum_j \hat{a}_j = K\,,
\]
we can write
\[
\sum_i b_i = \left( \sum_i (\Psi \hat{a})_i \right) \left( \sum_j e^{-(\Psi a)_j} \right) =\frac{1}{\| a \|_1} \left( \sum_i (\Psi a)_i \right) \left( \sum_j e^{-(\Psi a)_j} \right).
\]
Hence,
\begin{equation} \label{eq:b_sum_comp}
    \sum_i b_i = \frac{1}{\| a \|_1} \left( \sum_i x_i \right) \left( \sum_j e^{-x_j} \right).
\end{equation}
Substituting Eqns. \eqref{eq:b_bar_comp} and \eqref{eq:b_sum_comp} into Eq. \eqref{eq:KL_comp} yields
\[
\frac{d}{dt}(D_{\textrm{KL}}) = \frac{1}{(K-1)D \|a\|_1} \left[ (K-1) \sum_j x_j e^{-x_j}- \left( \sum_i x_i \right) \left( \sum_j e^{-x_j} \right) \right].
\]
Now, the bracketed term can be rewritten as 
$
\frac{1}{2} \sum_{ij} (x_i - x_j) \left( e^{-x_i} - e^{-x_j} \right),
$
yielding
\[
\frac{d}{dt}(D_{\textrm{KL}}) = \frac{1}{2(K-1)D \|a\|_1} \sum_{ij} (x_i - x_j) \left( e^{-x_i} - e^{-x_j} \right).
\]
For each pair $(i,j)$, the terms $x_i - x_j $ and $ e^{-x_i} - e^{-x_j}$ have opposite signs unless $x_i=x_j$. Hence, every summand is non-positive, vanishing only when $x_i=x_j$. Hence the full sum is always non-positive, and only equals zero if $x$ is a uniform vector. Because $D>0$ and $\|a\|_1>0$, the prefactor is strictly positive. Moreover, since $\Psi$ is invertible and has $1_{K-1}$ as an eigenvector, the condition $x \propto 1_{K-1}$ implies $a \propto 1_{K-1}$, and hence $\hat{a} = \frac{1}{K-1} 1_{K-1}$.

Thus $D_{\textrm{KL}}$ satisfies:

\begin{itemize}
    \item $D_{\textrm{KL}} \geq 0$, with equality only when $\hat{a} = \frac{1}{K-1} 1_{K-1}$.
    \item $\frac{d}{dt}(D_{\textrm{KL}}) \leq 0$, with equality only at $\hat{a} = \frac{1}{K-1} 1_{K-1}$.
\end{itemize}

Consequently, $D_{\textrm{KL}}$ decreases monotonically to its minimal value, guaranteeing convergence of $\hat{a}$ to $\frac{1}{K-1} 1_{K-1}$ as measured by this metric.

\section{Related Work}
\paragraph{Neural Collapse.} First identified in deep classification networks by Papyan et al. \cite{Papyan2020Prevalence}, NC has since become a central focus of theoretical analysis.Researchers have extensively studied the conditions under which NC emerges, primarily using the UFM \cite{Mixon2020, fang2021exploring} with explicit $L_2$ regularization. Within this framework, NC has been studied under a variety of loss functions \cite{Zhou2022, Mixon2020, ji2022an, Han2022}, consistently emerging as a globally optimal structure. Complementary landscape analyses have revealed favorable optimization properties, where the loss surface consists only of global minima and non-degenerate saddles \cite{zhou22c, zhu2021geometric, ji2022an}. Beyond the UFM, NC has also been investigated in other theoretical settings \cite{jacot2024wide, zangrando2024neural}. Extensions of NC research have explored its behavior in settings with a large number of classes \cite{jiang2023generalized} and imbalanced class distributions \cite{yang2022inducing, thrampoulidis2022imbalance, behnia2023implicit, fang2021exploring, Dang2023NeuralCI, hong2023neural}, where generalized forms of implicit geometries beyond the simplex ETF have been observed starting with \cite{thrampoulidis2022imbalance}. Generalizations from the hard label setting have also been considered for the multilabel setting \cite{li2023neural}, soft-labeled language setting \cite{Zhao2024ImplicitGO}, and multi-class regression \cite{andriopoulos2024prevalence}. The influence of dataset properties has also been explored \cite{hong2024beyond, kothapalli2024kernel}. Researchers have further asked whether NC extends beyond the final layer, with evidence coming both from empirical studies \cite{He2022, rangamani23, Parker2023NeuralCI} and from analyses of deep UFMs \cite{Tirer2022, Dang2023NeuralCI, garrod2024unifying, Sukenik2023, sukenik2024neural, Garrod2024ThePO}, as well as beyond the UFM framework altogether \cite{beaglehole2024average}.

NC has also been connected to a range of practical aspects of deep learning, including adversarial robustness \cite{su2023robustness}, transfer learning \cite{galanti2021role, li2022principled}, generalization \cite{hui2022limitations, Gao2023TowardsDT}, and out-of-distribution detection \cite{wu2024pursuing, zhang2024epa}. The impact of network architecture has been studied in ResNets \cite{li2024residual, Sukenik2025NeuralCI} and large language models (LLMs) \cite{wu2024linguistic, Zhao2025OnTG}. For a comprehensive overview of NC and its implications, see the survey by Kothapalli et al. \cite{Kothapalli2022}.

\paragraph{Implicit Bias.} A large body of work studies how gradient descent or gradient flow induces structure in the absence of explicit regularization. For logistic regression, Soudry et al. \cite{Soudry_2018} showed convergence to the max-margin solution. Ji et al. \cite{ji20a} related the implicitly regularized path to explicitly regularized solutions in binary classification, which had been earlier studied by Rosset et al. \cite{rosset2003margin}. Analyses of limiting behavior have since been extended to homogeneous \cite{lyu2019gradient,Ji_directional} and non-homogeneous networks \cite{cai2025implicit}, alternative optimizers \cite{gunasekar2018characterizing,azizan2018stochastic,sun2022mirror,pesme2024implicit,zhang2024implicit,fan2025implicit}, large step-sizes \cite{even2023s,wu2023implicit}, as well as, convolutional architectures \cite{Gunasekar2018ImplicitBO,lawrence2021implicit} and transformers \cite{ataee2023max,tarzanagh2023transformers,vasudevaimplicit,julistiono2024optimizing}. Many of these works focus on binary settings, where no interaction arises between different singular modes. This includes \cite{ji2018gradient}, which shows that the singular vectors of the parameter matrices will align. A complementary thread investigates implicit bias in linear networks with MSE loss, examining the effects of gradient descent and network depth \cite{pmlr-v80-arora18a, Yaras2023TheLO, Tu_mixed_dynam, bah2022learning}, including matrix factorization \cite{li2020towards, NEURIPS2019_c0c783b5, razin2020implicit, Gunasekar2017ImplicitRI}, and convergence properties \cite{arora2018convergence}. Owing to the simplicity of MSE, however, these analyses differ substantially from our setting.

Implicit bias has also been studied directly in the unregularized CE UFM. Ji et al. \cite{ji2022an} proved convergence to a KKT point of a constrained norm minimization problem. Thrampoulidis et al. \cite{thrampoulidis2022imbalance} characterized the solution found by the regularization path under imbalanced classes. Our results are the first to explicitly characterize the implicit bias of gradient descent dynamics in unregularized settings, and thus provide the first definitive link between implicit bias theory and neural collapse geometries.

\paragraph{Exact Dynamics.} Exact dynamics analyses study deep learning phenomena by reducing gradient flow to evolution equations for the singular values of the parameter matrices and solving these equations explicitly. To date, such works focus exclusively on linear networks trained with MSE loss. The approach was pioneered by Saxe et al. \cite{Saxe2013ExactST, Saxe2019} and extended by Gidel et al. \cite{Gidel2019} to weaker assumptions and to discrete gradient descent. These studies revealed that networks learn the hierarchical structure of singular values sequentially. Subsequent work generalized the technique to alternative initialization schemes \cite{Kunin2024GetRQ, Braun2022, tarmoun21a} and to investigations of the lazy-to-rich regime transition \cite{Braun2022, Domine2024FromLT}. Recent efforts have sought to broaden the scope beyond linear architectures. Mainali et al. \cite{Mainali2025ExactLD} adapted these methods to transformers, though with simplifications including linear attention mechanisms in place of softmax. Similarly, Zhao et al. \cite{zhao2025geometry} employed exact dynamics within the UFM to investigate semantic emergence from language data. However, all existing exact dynamics analyses—including these recent extensions—remain limited to MSE loss. Our work breaks this barrier by extending the framework to the CE loss, a more challenging yet practically essential setting.

\section{Conclusion}

In this work, we introduced a Hadamard Initialization scheme that preserves singular vector structure under the softmax operation, enabling an extension of the exact dynamics framework of Saxe et al. to the CE setting. Our analysis revealed striking differences from prior exact-dynamics results: the softmax nonlinearity couples the singular values, leading to complex dynamics, non-monotonic convergence, and exponential slowdowns. Such features are invisible to existing KKT-based analyses of the CE UFM. We further explained why such complex behaviors are rarely observed in practice, showing that standard initialization schemes naturally steer trajectories away from the regions where these effects manifest. Together, these findings provide the first path-level analysis of implicit bias in the unregularized CE UFM.

The Hadamard Initialization technique offers a tractable tool for studying CE dynamics and opens the door to extending exact-dynamics style analyses beyond linear MSE models. Natural next steps include examining how other results obtained via exact dynamics \cite{Gidel2019, Kunin2024GetRQ, Braun2022, tarmoun21a, Domine2024FromLT} can be recovered within our framework, as well as developing extensions to deep UFMs, settings with explicit regularization, and linear networks beyond the UFM construction. We hope this work motivates the use of these techniques to further investigate implicit bias and NC in more general, nonlinear architectures.

\section*{Acknowledgments}

CG is supported by the Charles Coulson Scholarship. The authors also acknowledge support from His Majesty’s Government in the development of this research. CT acknowledges generous support by NSERC. CT also thanks Yize Zhao for discussions on very early phases of this project. The authors are grateful to the organizers of the 2024 workshop ``Beyond classical regimes in statistical inference and machine learning'' in CIMI Toulouse where this collaboration was initiated. For the purpose of Open Access, the authors have applied a CC BY public copyright license to any Author Accepted Manuscript (AAM) version arising from this submission.

\newpage

\bibliographystyle{unsrt}  
\bibliography{references}  

\newpage
\appendix

\clearpage
\renewcommand{\contentsname}{Table of Contents} 
\tableofcontents
\clearpage

\section{Summary of Notations}
\label{sec:notations}
\begin{table}[h]
\centering
\caption{Summary of Key Notations}
\label{tab:notations}
\begin{tabular}{@{}lll@{}}
\toprule
\textbf{Notation} & \textbf{Description} & \textbf{Context / Definition} \\
\midrule
$K$ & Number of classes. & Assumed to be $K=2^m$. \\
$n$ & Number of samples per class. & Assumed to be balanced. \\
$W$ & Last-layer weight matrix. & $\R^{K \times d}$ \\
$H$ & Feature matrix. & $\R^{d \times Kn}$ \\
$Z$ & Logit matrix. & $Z = WH$ \\
$Y$ & One-hot label matrix. & $Y = I_K \otimes 1_n^T$ \\
$P(Z)$ & Softmax matrix. & Softmax applied columnwise to $Z$. \\
$S$ & Simplex ETF matrix. & $S = I_K - \frac{1}{K} 1_K 1_K^T$ \\
$\Phi$ & Sylvester Hadamard matrix. & $\R^{K \times K}$ matrix (Def. \ref{def:sylvester}). \\
$\Psi$ & Reduced dynamics matrix. & $1_{K-1} 1_{K-1}^T - \Phi[2:K,2:K]$. \\
$U, V$ & Left/Right singular vector matrices. & $U = V = \frac{1}{\sqrt{K}} \Phi$ for Hadamard Init. \\
$\alpha_i, \beta_i$ & Singular values. & Non-trivial singular values of $W$ and $H$. \\
$a$ & Logit singular values (vector). & $a = (a_1, \dots, a_{K-1})$, $a_i = \alpha_i \beta_i$. \\
$\hat{a}$ & Normalized logit singular values. & $\hat{a}_i = a_i / \|a\|_1$. \\
$b, D$ & Terms in reduced dynamics. & Defined in Eq. \ref{eq:b_and_D_def}. \\
$E(a)$ & Energy functional. & Reduced energy functional for $a$ (Eq. \ref{eq:energy_func}). \\
$M$ & $L_2$-based metric for NC. & Measures distance from $\hat{a}$ to uniform (Eq. \ref{eq:metric}). \\
$D_{\mathrm{KL}}$ & KL divergence. & Used as a Lyapunov function (Thm. \ref{thm:NC_conv}). \\
\bottomrule
\end{tabular}
\end{table}

Throughout the paper, we use $\R$ to denote the set of real numbers. $I_k$ represents the $k \times k$ identity matrix, and $1_k$ denotes an all-ones vector of dimension $k$. For a given matrix $A$, $A^T$ is its transpose, and $A[i_1:i_2, j_1:j_2]$ indicates the submatrix sliced from rows $i_1$ to $i_2$ and columns $j_1$ to $j_2$. The operator $\operatorname{diag}(v)$ forms a diagonal matrix using the elements of vector $v$. We use $\otimes$ for the Kronecker product, $\odot$ for the Hadamard (elementwise) product and $\oplus$ for the bitwise XOR. Finally, $\| \cdot \|_F$,  and $\| \cdot \|_1$ denote the Frobenius norm and the $L_1$ norm, respectively. Some key notation is summarized in Table \ref{tab:notations}.

\section{Further Theoretical Results} \label{sec:further_theory}

\subsection{Logit singular value dynamics as mirror flow}
Recall the dynamics of the logit singular values from Eq. \eqref{eq:logit_sing_evol} are given as
\begin{align}\label{eq:logit_sing_evol_appendix}
 \frac{da_i}{dt} = a_i\cdot\frac{\sum_{j=1}^{K-1} \Psi_{ij} e^{-(\Psi a)_j}}{1 + \sum_{j=1}^{K-1} e^{- (\Psi a)_j }}\,,\qquad i\in[K-1]\,.
\end{align}
Define the energy functional
\[
E(a)=\log \Big( 1+ \sum_{j=1}^{K-1} \exp \left( - (\Psi a)_j \right) \Big)\,.
\]
Then, the dynamics in Eq. \eqref{eq:logit_sing_evol_appendix} can be expressed conveniently in vector form as
\[
\dt{a} = -a\,\odot\, \nabla_a E(a)\,.
\]
For positive initialization, $a>0$ in the entire trajectory, and thus it holds equivalently that 
\[
\dt{\log(a)} = -\nabla_a E(a)\,.
\]
But, $\dt{\log(a)}=\dt{\nabla_a \phi(a)}$ where $\phi(a)=\sum_{i=1}^{K-1}a_i\log(a_i)$ is the negative entropy potential. Overall, we have expressed the dynamics of the logit singular values in the form of mirror-flow updates with respect to the entropy potential:
\[
\dt{\nabla_a\phi(a)} = -\nabla_a E(a)\,.
\]

\subsection{NC is the only full-rank stable direction}\label{sec:Nc only full rank}

Given the dynamics of Eq. \eqref{eq:normalized_dynamics}, if we look for a stable direction, meaning $\frac{d \hat{a}_i}{dt}=0$ for $i=1,...,K-1$, that also satisfies full-rank, meaning $\hat{a}_i >0$ for $i=1,\dots,K-1$, then this reduces to the condition:
\[
b_i = \bar{b}, \quad \textrm{for } i=1,...,K-1.
\]
This means that the vector $b$ can be written as $b = \bar{b} 1_{K-1}$, which imposes the following condition on the vector $a$
\[
\Psi e^{-(\Psi a)} = \bar{b} 1_{K-1}.
\]
Then using that $\Psi$ is invertible and $\Psi^{-1}1_{K-1} = \frac{1}{K} 1_{K-1}$ gives:
\[
a = - \frac{1}{K} \log \left( \frac{1}{K} \bar{b} \right) 1_{K-1},
\]
and if $a \propto 1_{K-1}$, it implies $\hat{a} = \frac{1}{K-1} 1_{K-1}$. Hence, the only full-rank stable direction is $\hat{a} = \frac{1}{K-1} 1_{K-1}$.

\subsection{Additional examples of non-monotonic behavior}\label{sec:examples non-monotonic}

\subsubsection{Reverse-KL metric}

One can consider the reverse-KL divergence, defined as
\[
D_{\mathrm{KL}} \left(\hat{a} \,\|\, \frac{1}{K-1} 1_{K-1} \right)= \log(K-1) + \sum_i \hat{a}_i \log(\hat{a}_i).
\]
We will denote it by $\tilde{D}_{\textrm{KL}}$ for this section so as to differentiate it from the KL divergence of the main text. It has derivative
\[
\frac{d \tilde{D}_{\textrm{KL}}}{dt} = \frac{1}{D} \sum_i [b_i - \bar{b}] \hat{a}_i \log( \hat{a}_i).
\]
In $K=8$, a counter-example, demonstrating $\tilde{D}_{\textrm{KL}}$ is not monotonic decreasing, occurs for 
$$a=[1,1,1,1.25,0.0001,0.0001,0.0001],$$
where $\frac{d \tilde{D}_{\textrm{KL}}}{dt} \approx 0.0037$.

\subsubsection{Frobenius Distance}

Typically in other works, the rate of convergence to NC is measured through the following metric:
\[
\tilde{M} = \left\| \frac{Z}{\| Z \|_F} - \frac{S}{\| S\|_F} \right\|_F^2, 
\]
which, when expressed in terms of our logit singular values (after absorbing constants as in Eq. \eqref{eq:logit_sing_evol}), is given by
\[
\tilde{M} = \sum_{i=1}^{K-1} \left[ \frac{a_i}{\sqrt{\sum_j a_j^2}} - \frac{1}{\sqrt{K-1}} \right]^2 .
\]
In the main text, to avoid introducing multiple normalizations, we normalized with the $L_1$ norm, but the metric $\tilde{M}$ displays the same behavior as the metric $M$. First note that its derivative is given by
\[
\frac{d \tilde{M}}{dt}=\frac{2}{D\| Z\|_F^3 \sqrt{K-1}} \left[ \sum_{ij} a_i a_j b_j (a_j - a_i) \right],
\]
where we continue to denote $\sqrt{\sum_j a_j} = \| Z\|_F$. For this to be non-positive we require that:
\[
\sum_{ij} a_i a_j b_j [a_j - a_i] \leq 0 .
\]
A counter-example occurs for $a=[1,1,1,1.25,0.0001,0.0001,0.0001]$, where the LHS of the above equality is $\approx 0.088$. Note the existence of this counter-example for $K=8$ is sufficient for counter-examples to exist for all $K \geq 8$, using the same reasoning as provided for the proof of Theorem \ref{thm:non_monotone} in Appendix \ref{sec:non_monotone_proof}.

\subsubsection{Log-ratio of max-min singular values}

Define the following metric:
\[
M_{\textrm{LR}} = \max_i (\log(a_i)) - \min_j(\log(a_j)).
\]
This quantity is monotonic decreasing for $K \leq 8$, but non-monotonic for $K \geq 16$.

We first prove this quantity is monotonic decreasing for $K=8$:

\begin{proposition}
    Let $K=8$, and $a(t) \in \mathbb{R}^{K-1}$ be a variable initialized with $a_i(0)>0$ for all $i$, that evolves under the time evolution of Eq. \eqref{eq:logit_sing_evol}. Then the metric $M_{\textrm{LR}}$, is monotonic decreasing.
\end{proposition}

\textbf{Proof:}

Recall our dynamics, given in Eq. \eqref{eq:logit_sing_evol}. By multiplying the numerator and denominator by  $\exp( 2 \sum_j a_j)$, we can rewrite this as 
\[
\frac{d}{dt}(a_i)=\frac{1}{D} b_i a_i,
\]
where
\[
b_i = \sum_j \Psi_{ij} \exp \left( (\left[2 (1_{K-1} 1_{K-1}^T) - \Psi \right]a)_j \right), \quad D= \exp( 2 \sum_j a_j) + \sum_j \exp \left( (\left[2 (1_{K-1} 1_{K-1}^T) - \Psi \right]a)_j \right).
\]
Expanding this out directly produces the following expressions for the variables $b_i$
$$b_1 = \exp(2(a_2+a_4+a_6)) + \exp(2(a_3+a_4+a_7))+\exp(2(a_2 + a_5 + a_7)) + \exp(2(a_3 + a_5 + a_6))$$
$$b_2 = \exp(2(a_1+a_4+a_5)) + \exp(2(a_3+a_4+a_7))+\exp(2(a_1 + a_6 + a_7)) + \exp(2(a_3 + a_5 + a_6))$$
$$b_3 = \exp(2(a_2+a_4+a_6)) + \exp(2(a_1+a_4+a_5))+\exp(2(a_2 + a_5 + a_7)) + \exp(2(a_1 + a_6 + a_7))$$
$$b_4 = \exp(2(a_1+a_2+a_3)) + \exp(2(a_2+a_5+a_7))+\exp(2(a_1 + a_6 + a_7)) + \exp(2(a_3 + a_5 + a_6))$$
$$b_5 = \exp(2(a_2+a_4+a_6)) + \exp(2(a_3+a_4+a_7))+\exp(2(a_1 + a_2 + a_3)) + \exp(2(a_1 + a_6 + a_7))$$
$$b_6 = \exp(2(a_1+a_4+a_5)) + \exp(2(a_3+a_4+a_7))+\exp(2(a_1 + a_2 + a_3)) + \exp(2(a_2 + a_5 + a_7))$$
$$b_7 = \exp(2(a_2+a_4+a_6)) + \exp(2(a_1+a_4+a_5))+\exp(2(a_1 + a_2 + a_3)) + \exp(2(a_3 + a_5 + a_6))$$
From this one can note the following properties immediately:

\begin{enumerate}
    \item There are seven triples of the singular values $a_i$ that appear in the exponents.
    \item The triples that feature $a_i$ do not appear in the expression of $b_i$.
    \item Each $a_i$ appears with every other $a_j$ in exactly one triple.
    \item Every pair of $b_i,b_j$ share exactly two terms, and differ in two. The two they share are precisely the two not featuring either of their corresponding $a_i$.
\end{enumerate}

Now suppose that at some time $t$ we have that the largest mode is given by $a_i$ and the smallest by $a_j$, then the log ratio of these singular values has derivative 
given by
\[
\frac{d}{dt} \left( \log (a_i) - \log(a_j) \right) = \frac{1}{D} [b_i - b_j].
\]
Because of the properties of the expressions for the variables $b_i$ described above, the difference $b_i - b_j$ takes the following form:
\[
b_i - b_j = x_j x_p x_q + x_j x_r x_s - x_i x_r x_p - x_i x_s x_q,
\]
where we have defined $x_i = \exp(2a_i)$ for the purpose of this section. We can alternatively write this expression as
\[
b_i - b_j = x_p[x_j x_q - x_i x_r] + x_s [ x_j x_r - x_i x_q],
\]
and further as
\[
b_i - b_j = x_p x_i x_q \left[ \frac{x_j}{x_i} - \frac{x_r}{x_q} \right]+ x_s x_i x_r \left[ \frac{x_j}{x_i} - \frac{x_q}{x_r} \right].
\]
Now note that, because $x_j$ is the smallest, and $x_i$ is the largest, we must have:
\[
x_j \leq \textrm{any other choice of } x, \qquad \frac{1}{x_i} \leq \frac{1}{\textrm{any other choice of }x}.
\]
This implies that:
\[
\frac{x_j}{x_i} \leq \frac{x_r}{x_q}, \qquad \frac{x_j}{x_i} \leq \frac{x_q}{x_r},
\]
and hence the difference $b_i - b_j$ is non-positive, and only takes value zero when
\[
\frac{x_j}{x_i} = \frac{x_r}{x_q}= \frac{x_q}{x_r},
\]
but this can only happen if $x_j=x_i$, meaning the largest and smallest mode must be equal, which implies $\hat{a} = \frac{1}{K-1} 1_{K-1}$. Hence the quantity $M_{\textrm{LR}}$ always has a negative derivative, except when we have a neural collapsed solution.

$\Box$

It remains to show a counter-example in $K=16$. Let $a = [5.01,0.99,2,5,1,5,1,2,1,5,3,5,1,5,1]$, then the largest singular value is $a_1$, and the smallest is $a_2$, but 

$$\frac{d}{dt} \left( \log(a_1) - \log(a_2) \right) \approx 1.8 \times 10^{-13}$$

Hence this metric is not monotonic decreasing for $K=16$.

\section{Non-uniform convergence points} \label{sec:non_uniform_conv_points}

 \begin{figure}
     \centering
     \begin{subfigure}{0.33\textwidth}
         \centering
         \includegraphics[width=\textwidth]{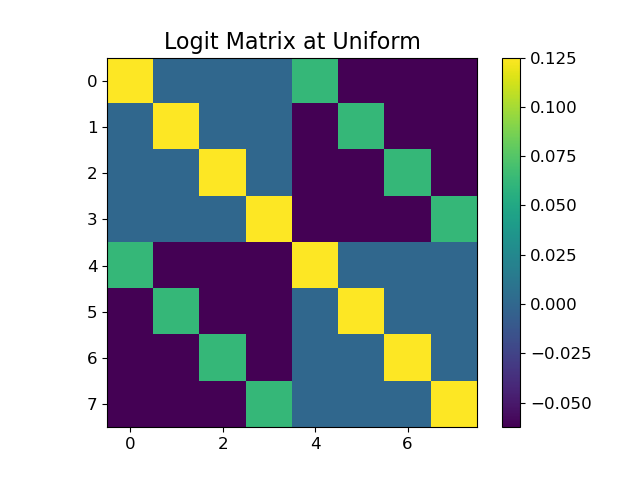}
     \end{subfigure}
     \begin{subfigure}{0.33\textwidth}
         \centering
         \includegraphics[width=\textwidth]{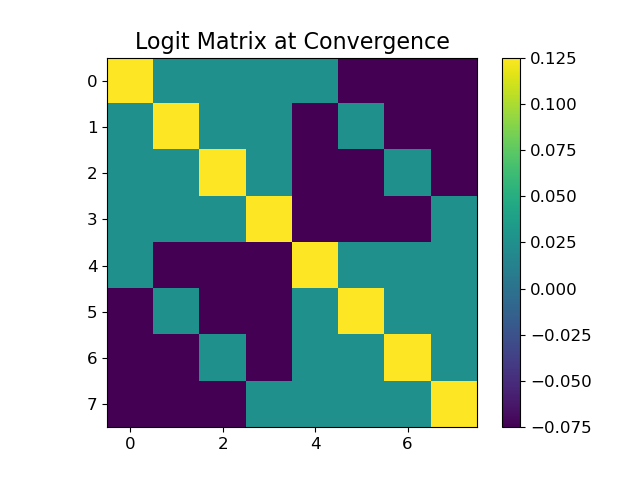}
     \end{subfigure}
     \caption{Plots of the normalized logit matrices generated by the vector $a^* = [1,1,1,1,0,0,0]$, which is uniform on its support, and by the non-uniform vector $a_{\textrm{conv}}=[1,1,1,2,0,0,0]$. The smallest normalized margin achieved by $a_{\textrm{conv}}$ is visibly larger, indicating that this configuration outperforms the uniform one as the logit norm diverges.}
     \label{fig:imbalanced_logits}
 \end{figure}

Here we provide an expanded discussion of the example shown in Figure \ref{fig:non_monotone_conv}, in order to build intuition for why non-uniform convergence points can arise in the loss surface for general $K$. Figure \ref{fig:imbalanced_logits} displays the normalized logit matrix with singular-values vector $a^*=[1,1,1,1,0,0,0]$ that lies very close to the initialization used in Figure \ref{fig:non_monotone_conv}. Since the set of indices for which $a_i \neq 0$ remains stable under the flow, we might expect that the convergence point reached from  initialization $a(0)=a^*$ is the minimizer among all configurations with at most the first four indices nonzero.

It is straightforward to show that the following quantity decreases monotonically along the flow:
$$E(a)=\log \Big( 1+ \sum_{j=1}^{K-1} \exp \left( -\|a\|_1 (\Psi \hat{a})_j \right) \Big)\,,$$
where we have chosen to express it in terms of $\hat{a}$ and $\| a \|_1$. This coincides with the energy functional of Eq. \eqref{eq:energy_func} up to absorbed constants. In particular, as $\|a\|_1 \to \infty$, the energy is asymptotically:
$$E \simeq \sum_{j=\textrm{argmin}_u \{ (\Psi \hat{a})_u\}} \exp \left( -\|a\|_1 (\Psi \hat{a})_j \right) \Big)\,.$$
Thus, the smallest normalized margin $(\Psi \hat{a})_i$ dominates the energy, and we therefore expect the flow as $t \to \infty$ to maximize this quantity. The normalized margins correspond to the difference between the diagonal and off-diagonal entries of the normalized logit matrix. 

Comparing the normalized logit matrix at the uniform configuration $a^*$ with that of the configuration $a_{\textrm{conv}}$, as shown in Figure \ref{fig:imbalanced_logits}, we see that $a^*$ has a smaller minimal margin.

Numerically, what occurs in Figure \ref{fig:non_monotone_conv} of the main text is that the last three singular values are initially so small that the dynamics treats them effectively as zero. The remaining singular values, therefore, evolve toward the minimizer supported on only the first four indices. This minimizer is not the uniform point, as indicated by its worse margin in Figure \ref{fig:imbalanced_logits}; instead, it is the configuration $a_{\textrm{conv}}$,in which one singular value is larger than the others. This is the structure toward which the flow in Figure \ref{fig:non_monotone_conv} trends over the considered time interval. Because the flow approaches a non-uniform point, most common measures of uniformity increase, and thus fail to act as Lyapunov functions for the dynamics. The KL divergence avoids this issue by effectively accounting for the growth of the three small singular values, which offsets the non-uniform behavior of the remaining ones.

\section{Proofs}

\subsection{Proof of Theorem \ref{thm:sing_val_evol}} \label{sec:sing_bal_evol_proof}

Recall $U= \frac{1}{\sqrt{K}} \Phi$, where $\Phi$ is the $K \times K$ Sylvester Hadamard matrix, and that we initialize our parameter matrices as: $W=UD_W R^T$, $H=R D_H V^T$, where $R \in \mathbb{R}^{d \times d}$ is orthogonal, and $V=U \otimes Q$ is orthogonal, with $Q$ being a right singular matrix of $1_n^T$, meaning it obeys the relation: $1_n^T = \sqrt{n} e_1^{(n)T} Q^T$, where $e_1^{(n)}$ is the first standard basis vector in $\mathbb{R}^n$. Also recall that $D_W \in \mathbb{R}^{K \times d}$ has its only non-zero singular values given by $\alpha_1,\dots \alpha_K$, and $D_H \in \mathbb{R}^{d \times Kn}$ has its only non-zero singular values given by $\beta_1,\dots, \beta_K$. Note as a consequence of the definitions of $U,V$, we can write $Y=\sqrt{n}U[I_K,0_{K \times K(n-1)}]V^T$.

We also have at initialization that the logit matrix is given by
\[
Z=UD_W D_HV^T.
\]
Note that $D_W \in \mathbb{R}^{K \times d}$ and $D_H \in \mathbb{R}^{d \times Kn}$. Hence, defining $D_Z=D_W D_H \in \mathbb{R}^{K \times Kn}$, and $D_a = \textrm{diag}(a_1,...,a_{K})$, where $a_i = \alpha_i \beta_i$, this matrix can be written as:
\[
D_Z = [D_a,0_{K \times K(n-1)}] = D_a \otimes e_1^{(n)T}.
\]
Hence we can write 
\[
Z=U \left( D_a \otimes e_1^{(n)T} \right)(U \otimes Q)^T = U \left[ (D_a U^T) \otimes (e_1^{(n)T} Q^T) \right].
\]
Next, using that $1_n^T = \sqrt{n} e_1^{(n)T} Q^T$, this becomes:
\[
Z = \frac{1}{\sqrt{n}} U\left[ (D_a U^T) \otimes 1_n^T \right].
\]
Then using that this has repeated columns, this becomes:
\[
Z =  (UD_{\tilde{a}}U^T) \otimes 1_n^T,
\]
where we define $\tilde{a}=\frac{1}{\sqrt{n}}a$. Since this is the same matrix as $U D_{\tilde{a}} U^T$, just with repeated columns, we can apply Theorem \ref{thm:mutual_diag} to calculate the softmax matrix at initialization:
\[
P(Z)=(UD_{\tilde{\nu}}U^T) \otimes 1_n^T,
\]
where
\[
\tilde{\nu}_i= \frac{\sum_j \Phi_{ij} e^{\frac{1}{K} (\Phi \tilde{a})_j}}{\sum_j e^{\frac{1}{K}(\Phi \tilde{a})_j}}.
\]
Now consider the gradient flow equations of our model
\[
\frac{dW}{dt} = (Y-P)H^T, \qquad \frac{dH}{dt}=W^T (Y-P).
\]
Consider the change of variable given by $W=U \tilde{W}$, $H=\tilde{H}V^T$, these have evolution equations given by:
\[
\frac{d}{dt}(\tilde{W})=U^T(Y-P)V \tilde{H}^T, \qquad \frac{d}{dt}(\tilde{H})=\tilde{W}^T U^T (Y-P)V.
\]
Note that, by construction, both the matrices $Y,P$ can be written in SVD form through the matrices $U,V$
\[
Y = \sqrt{n} U D_Y V^T, \qquad P=\sqrt{n}UD_P V^T,
\]
where $D_Y =[I_K,0_{K \times K(n-1)}] \in \mathbb{R}^{K \times Kn}$ and $D_P = [\textrm{diag}(\tilde{\nu}),0_{K \times K(n-1)}] \in \mathbb{R}^{K \times Kn}$, and so
\[
U^T(Y-P)V  = \sqrt{n}[\textrm{diag}(1_K-\tilde{\nu}),0_{K \times K(n-1)}],
\]
label this matrix as $B$. Now returning to the evolution equations for $\tilde{H}$ and $\tilde{W}$, we label the rows of $\tilde{W}$ by $\tilde{w}_a$, and the columns of $\tilde{H}$ by $\tilde{h}_a$. These have evolution equations given by:
\[
\frac{d}{dt}((\tilde{w}_i)_j) = \sum_x B_{ix} (\tilde{h}_x)_j, \qquad \frac{d}{dt}((\tilde{h}_j)_i) = \sum_x B_{xj} (\tilde{w}_x)_i.
\]
Now using that $B_{ix}=\sqrt{n} (1 - \tilde{\nu}_i) \delta_{ix}$ gives:
\begin{equation} \label{eq:saxe_construct}
    \frac{d}{dt}((\tilde{w}_i)_j) = \sqrt{n} (1- \tilde{\nu}_i) (\tilde{h}_i)_j, \qquad \frac{d}{dt}((\tilde{h}_j)_i) = \sqrt{n} (1-\tilde{\nu}_j)(\tilde{w}_j)_i,
\end{equation}
where in the second of these equations if $j > K$ the derivative is zero.

Note that using our initialization scheme, we have at initialization $\tilde{w}_i = \alpha_i r_i$, $\tilde{h}_i = \beta_i r_i$, for $i=1,\dots,K$, and 0 otherwise, where $r_i$ are the columns of $R$. One observes from Eq. \eqref{eq:saxe_construct} that the derivatives also point in the same direction. Hence this property is maintained at all times, meaning the vectors $\tilde{w}_i, \tilde{h}_i$ only evolve in their coefficients, not in their directions. These coefficients then evolve as:
\[
\frac{d}{dt}(\alpha_i) = \sqrt{n} (1- \tilde{\nu}_i) \beta_i, \qquad \frac{d}{dt}(\beta_i) = \sqrt{n} (1-\tilde{\nu}_i)\alpha_i.
\]
Translating back to our original parameter matrices $W,H$, these are the evolution equations for their singular values, and the singular vectors are frozen for all times.

It remains to demonstrate these equations take the exact form in the theorem statement. Note from the expression of $\tilde{\nu}_i$ we have
\[
1-\tilde{\nu}_i = \frac{1}{\sum_j e^{\frac{1}{K}(\Phi \tilde{a})_j}} \left[ \sum_j (1-\Phi_{ij}) e^{\frac{1}{K} (\Phi \tilde{a})_j} \right].
\]
Then dividing the numerator and denominator by $\exp(\frac{1}{K} \sum_j \tilde{a}_j)$, gives:
\[
1-\tilde{\nu}_i = \frac{1}{\sum_j e^{-\frac{1}{K}([1_K 1_K^T - \Phi] \tilde{a})_j}} \left[ \sum_j [1_K 1_K^T - \Phi]_{ij} e^{-\frac{1}{K} ([1_K 1_K^T - \Phi] \tilde{a})_j} \right]=\frac{b_i}{D},
\]
and hence this gives the desired form, once we recall that $\tilde{a}_i = \frac{1}{\sqrt{n}} a_i$.

\subsection{Proof of Theorem \ref{thm:non_monotone}} \label{sec:non_monotone_proof}

Recall our choice of metric is given by:

$$M= \frac{1}{2} \sum_{i=1}^{K-1} \Big( \hat{a}_i - \frac{1}{K-1} \Big)^2\,.$$

A simple calculation, using the derivatives provided in Eq. \eqref{eq:normalized_dynamics}, gives 

\begin{equation} \label{eq:metric_deriv}
    \frac{dM}{dt} = \frac{1}{D} \sum_i \hat{a}_i^2[b_i - \bar{b}]\,.
\end{equation}

First note that $M=0$ in the $K=2$ case for all positive initializations, and so trivially is monotonic. We begin by proving monotonicity for $K=4$.

Note in $K=4$ we have
\[
\Psi = \begin{bmatrix}
    2 & 0 & 2 \\
    0 & 2 & 2 \\
    2 & 2 & 0
\end{bmatrix} \implies \Psi a = \begin{bmatrix}
    2a_1 + 2a_3 \\
    2 a_2 + 2a_3 \\
    2a_1 + 2a_2
\end{bmatrix} \implies \exp(- \Psi a) = \begin{bmatrix}
    e^{-2a_1 - 2a_3} \\
    e^{-2a_2 - 2a_3} \\
    e^{-2a_1 - 2a_2}
\end{bmatrix}.
\]
Hence
\[
\frac{1}{D}b = \frac{2}{1+ e^{-2a_1 - 2a_3} + e^{-2a_2 - 2a_3} + e^{-2a_1 - 2a_2}} \begin{bmatrix}
    e^{-2a_1 - 2a_3} + e^{-2a_1 - 2a_2} \\
    e^{-2a_2 - 2a_3} + e^{-2a_1 - 2a_2} \\
    e^{-2a_1 - 2a_3} + e^{-2a_2 - 2a_3}
\end{bmatrix}.
\]
Multiplying numerator and denominator by $\exp(2 \sum_i a_i)$ gives
\[
\frac{1}{D}b = \frac{2}{e^{2(a_1 + a_2 + a_3)} + e^{2 a_1} + e^{2 a_2} + e^{2 a_3}} \begin{bmatrix}
    e^{2a_2} + e^{2 a_3} \\
    e^{2a_1} + e^{2 a_3} \\
    e^{2a_1} + e^{2 a_2}
\end{bmatrix}.
\]
Writing $x_i = \exp(2a_i)$, this is then
\[
\frac{1}{D}b = \frac{2}{x_1x_2x_3 + \sum_i x_i} \begin{bmatrix}
    \sum_i x_i-x_1 \\
    \sum_i x_i-x_2 \\
    \sum_i x_i-x_3
\end{bmatrix},
\]
and then $\bar{b}$ is given by:
\[
\frac{1}{D} \bar{b} = \frac{2}{x_1x_2x_3 + \sum_i x_i} [ \sum_i x_i - \hat{a} \cdot x ].
\]
Hence using the expression for the derivative of our metric, given in Eq. \eqref{eq:metric_deriv}, we arrive at:
\[
\frac{dM}{dt} = \frac{2}{x_1x_2x_3 + \sum_i x_i} \sum_i \hat{a}_i^2 [ \hat{a} \cdot x - x_i].
\]
The coefficient is strictly positive, and so to argue non-increasing we can drop it. Hence we simply need to show:
\[
(\hat{a} \cdot x) \left( \sum_i \hat{a}_i^2 \right) - \sum_i \hat{a}_i^2x_i \leq 0,
\]
or equivalently:
\[
\sum_i \hat{a}_i^2 \leq \frac{\sum_i \hat{a}_i^2x_i}{\sum_i \hat{a}_i x_i}.
\]
Define the quantities:
\[
q_i = \frac{\hat{a}_ix_i}{\sum_i \hat{a}_i x_i},
\]
noting that they form a probability distribution. Then our inequality can be viewed as comparing the expectation of $\hat{a}$ over two different probability distributions, one with probabilities $\hat{a}$ and the other with probabilities $q$.

It is then simple to note that, since for any $\|a\|_1 >0$, $x_i$ is an increasing function of $\hat{a}_i$, the second distribution places more weight on larger values of $\hat{a}$, and so must be at least as large as the other expectation. They are only equal when there are no larger values of $\hat{a}_i$, meaning $\hat{a} = \frac{1}{K-1} 1_{K-1}$. Hence the inequality is satisfied, and the metric is monotonically decreasing.

It remains to show that the metric is not necessarily monotonic decreasing for $K \geq 8$. To show this for $K=8$, it is sufficient to note that for $a=[1,1,1,1.25,0.01,0.01,0.01]$ the derivative of our metric takes value $\frac{dM}{dt} \approx7.3 \times 10^{-4}$, and so the metric is increasing for these values. Call this specific counter-example $v^*$, and denote its corresponding $b$ vector and its weighted average, by $b^*$ and $\bar{b}^*$.

For $K>8$, define our initialization as $a=[v^*,0_{K-8}]$, then the derivative of our metric becomes 
\[
\frac{dM}{dt} =\frac{1}{D} \sum_{i=1}^7 (\hat{v}_i^*)^2 [b_i - \bar{b}], \qquad \textrm{where } \bar{b} = \sum_{j=1}^7 v_i^* b_i.
\]
Note that only $b_1,...,b_7$ are necessary for the computation of the sign. We now use the lemma stated and proven at the end of this section, which gives that when $a$ is only non-zero on at most the first $\frac{K}{2}-1$ entries, then the first $\frac{K}{2}-1$ entries of the $b$ vector are just equal to double the values they would be in the $\frac{K}{2}$ dimensional case. Applying this iteratively, we get that if $K=2^m>8$, then
\[
b_i = 2^{m-3} b^*_i, \qquad \bar{b}=2^{m-3} \bar{b}^*, \qquad \textrm{for } i=1,...,7,
\]
and so
\[
\frac{dM}{dt} =\frac{2^{m-3}}{D} \sum_{i=1}^7 (\hat{v}_i^*)^2 [b^*_i - \bar{b}^*], \qquad \textrm{where } \bar{b} = \sum_{j=1}^7 v_i^* b_i,
\]
but this clearly has the same sign as the counter-example in $K=8$, and so gives a positive derivative.

This provides a counter-example in any $K$, but it has entries that are zero, which is not permitted. However if we consider the vector
\[
a=[v^*, \epsilon 1_{K-8}],
\]
then by continuity, as $\epsilon \to 0$ this tends to the positive value achieved by $a=[v^*, 0_{K-8}]$. Hence for any $K$ there exists a non-zero value of $\epsilon$ such that $\frac{dM}{dt}>0$ for $a=[v^*, \epsilon 1_{K-8}]$. Hence counter-examples exist in all $K \geq 8$. 

It only remains to prove the following Lemma. 
\begin{lemma} \label{lem:counter_example}
    Let $K=2^m$, and $a^* \in \mathbb{R}^{K-1}$, with $b(a^*)=b^* \in \mathbb{R}^{K-1}$ and $\bar{b}^* = \sum_i \hat{a}_i^* b_i^* \in \mathbb{R}^{K-1}$. Let $a' \in \mathbb{R}^{2K-1}$ be given by $a' = [a^*,0_K]$, and consider the evolution under the dynamics detailed in Eq. \eqref{eq:logit_sing_evol}, but for dimension $2K$, initialized at $a'$. Denote the corresponding vectors $b(a'),\bar{b}(a') \in \mathbb{R}^{2K-1}$ by $b'$ and $\bar{b}'$ respectively. Then for $i=1,...,K-1$, we have
    \[
    b'_i=2b^*_i, \qquad \bar{b}' = 2 \bar{b}^*.
    \]
\end{lemma}

\textbf{Proof of Lemma \ref{lem:counter_example}: }

Let $\Psi_{K-1}$ and $\Psi_{2K-1}$ denote the two matrices derived from the Hadamard matrix, as defined in Sec. \ref{sec:hadamard_initialization}, with the subscript denoting their dimension. Using the construction in terms of Sylvester Hadamard matrices, as defined in Definition \ref{def:sylvester}, these are related in the following way:
\[
\Psi_{2K-1} = \begin{bmatrix}
    \Psi_{K-1} & 0_{K-1} & \Psi_{K-1} \\
    0_{K-1}^T & 2 & 21_{K-1}^T \\
    \Psi_{K-1} & 21_{K-1} & 21_{K-1} 1_{K-1}^T - \Psi_{K-1}
\end{bmatrix}.
\]
Hence
\[
\Psi_{2K-1} a' = \begin{bmatrix}
    \Psi_{K-1} & 0_{K-1} & \Psi_{K-1} \\
    0_{K-1}^T & 2 & 21_{K-1}^T \\
    \Psi_{K-1} & 21_{K-1} & 21_{K-1} 1_{K-1}^T - \Psi_{K-1}
\end{bmatrix} \begin{bmatrix}
    a^* \\
    0 \\
    0_{K-1}
\end{bmatrix} = \begin{bmatrix}
    \Psi_{K-1} a^* \\
    0 \\
    \Psi_{K-1} a^*
\end{bmatrix},
\]
and so 
\[
\exp(-\Psi_{2K-1} a) = \begin{bmatrix}
    \exp(-\Psi_{K-1} a^*) \\
    1 \\
    \exp(-\Psi_{K-1} a^*) 
\end{bmatrix},
\]
giving
\[
b' = \Psi_{2K-1} \exp(-\Psi_{2K-1} a) = \begin{bmatrix}
    \Psi_{K-1} & 0_{K-1} & \Psi_{K-1} \\
    0_{K-1}^T & 2 & 21_{K-1}^T \\
    \Psi_{K-1} & 21_{K-1} & 21_{K-1} 1_{K-1}^T - \Psi_{K-1}
\end{bmatrix} \begin{bmatrix}
    \exp(-\Psi_{K-1} a^*) \\
    1 \\
    \exp(-\Psi_{K-1} a^*) 
\end{bmatrix}.
\]
From this it is clear that the top $K-1$ entries of $b'$ are given by the vector $2\Psi_{K-1} \exp(-\Psi_{K-1} a) = 2 b^*$. Also note
\[
\bar{b}' = \sum_{j} \hat{a}'_j b_j' = \sum_{j=1}^{K-1} \hat{a}_j^* b_j' = 2 \bar{b}^*.
\]

\subsection{Proof of Theorem \ref{thm:linearized}} \label{sec:linearized_proof}

Recall our dynamics are given by:
\[
\frac{da_i}{dt} = \frac{1}{D} a_i b_i, \qquad \textrm{where: } b_i = \sum_j \Psi_{ij} e^{- (\Psi a)_j}, \quad D=1+ \sum_j e^{- (\Psi a)_j}.
\]
Suppose that $\|a\|_1 = \epsilon \ll 1$. Then expanding in $\epsilon$ gives $e^{- (\Psi a)_j} = 1 - (\Psi a)_j + O(\epsilon^2)$. Also recall that $\sum_j \Psi_{ij} = K$. We hence have:
\[
D = 1 + \sum_j [1 - (\Psi a)_j] + O( \epsilon^2)
\]
\[
= K -K \sum_j a_j + O( \epsilon^2)
\]
and hence:
\[
D = K(1- \epsilon) + O( \epsilon^2).
\]
Similarly we have:
\[
b_i = \sum_j \Psi_{ij} (1 - (\Psi a)_j) + O( \epsilon^2)
\]
\[
= K - (\Psi^2 a)_i + O ( \epsilon^2).
\]
Note, using the properties of the original underlying Hadamard matrix, we have $\Psi^2 = K(I+ 1_{K-1} 1_{K-1}^T)$, giving
\[
b_i = K - K a_i -K \epsilon + O( \epsilon^2).
\]
Hence, our differential equation for $a_i$ is given by:
\[
\frac{da_i}{dt} = a_i \left[K - K a_i -K \epsilon + O( \epsilon^2) \right] \left[ K(1- \epsilon) + O( \epsilon^2) \right]^{-1}
\]
\[
= a_i \left[1 - a_i - \epsilon + O( \epsilon^2) \right] \left[ 1- \epsilon + O( \epsilon^2) \right]^{-1}
\]
\[
 = a_i \left[1 - a_i - \epsilon + O( \epsilon^2) \right] \left[ 1+ \epsilon + O( \epsilon^2) \right]
\]
\[
 = a_i (1 - a_i) + O(\epsilon^3).
\]
Hence, we see that the linear dynamics do decouple. This equation is simple to solve, and gives a sigmoid solution, specifically:
\[
a_i(t) = \frac{a_i (0)}{a_i (0) + (1-a_i(0))e^{-t}}.
\]
This solution is only valid as an approximation in the linearized regime.

\subsection{Proof of Theorem \ref{thm:logit_case}} \label{sec:logit_proof}

When optimizing the logits directly, as detailed in Eq. \eqref{eq:logit_case}, the gradient flow equations are given by:
\[
\frac{dZ}{dt} = Y-P.
\]
We initialize our matrix $Z$ as $Z=U D_Z V^T$, where $U=\frac{1}{\sqrt{K}} \Phi$, and $V= U \otimes Q$, where $Q$ is a right singular matrix of $1_n^T$, meaning it obeys $1_n^T = \sqrt{n} e_1^{(n)T} Q^T$, where $e_1^{(n)}$ is the first standard basis vector in $\mathbb{R}^n$. Also recall $D_Z$ has its only non-zero singular values given by $a_1,\dots,a_K$. The exact same arguments used in the proof of Theorem \ref{thm:sing_val_evol} then gives that, at initialization, $Z$ can be written as
\[
Z = (U D_{\tilde{a}}U^T) \otimes 1_n^T,
\]
where $D_{\tilde{a}} = \textrm{diag}(\tilde{a}_1,\dots, \tilde{a}_K)$, with $\tilde{a}_i = \frac{1}{\sqrt{n}} a_i$. Then Theorem \ref{thm:mutual_diag} gives us that
\[
P(Z) = (U D_{\tilde{\nu}} U^T) \otimes 1_n^T,
\]
where
\[
\tilde{\nu}_i= \frac{\sum_j \Phi_{ij} e^{\frac{1}{K} (\Phi \tilde{a})_j}}{\sum_j e^{\frac{1}{K}(\Phi \tilde{a})_j}}.
\]
We can now turn to our gradient flow equation, and use the change of variable $Z = U \tilde{Z} V^T$, which evolves as:
\[
\frac{d}{dt} \left( \tilde{Z} \right) = U^T (Y-P) V.
\]
As in Theorem \ref{thm:sing_val_evol}, define $B=U^T (Y-P)V$. Since we have $Y= \sqrt{n} U D_Y V^T$ and $P = \sqrt{n} U D_P V^T$, where $D_Y =[I_K,0_{K \times K(n-1)}] \in \mathbb{R}^{K \times Kn}$ and $D_P = [\textrm{diag}(\tilde{\nu}),0_{K \times K(n-1)}] \in \mathbb{R}^{K \times Kn}$, we have:
\[
B = \sqrt{n}[\textrm{diag}(1_K-\tilde{\nu}),0_{K \times K(n-1)}].
\]
Hence the evolution in $\tilde{Z}$ becomes:
\[
\frac{d}{dt} \left( \tilde{Z}_{ij} \right) = \sqrt{n} (1-\tilde{\nu}_i) \delta_{ij}.
\]
We see that the only entries of $\tilde{Z}$ that evolve are the entries on its diagonal. Consequently if at initialization it takes the form of a singular value matrix, this property continues to hold for all future times. At initialization we have $\tilde{Z} = [\textrm{diag}(a),0_{K \times K(n-1)}]$, and so these singular values evolve as:
\[
\frac{d}{dt} (a_i) = \sqrt{n} (1 - \tilde{\nu}_{i}).
\]
Translating to the original variable $Z$, this tells us that the singular vectors remain fixed, and all time evolution occurs in the singular values.

Just as in the proof of Theorem \ref{thm:sing_val_evol}, using $\tilde{a} = \frac{1}{\sqrt{n}} a$ we can write:
\[
(1 - \tilde{\nu}_{i}) = \frac{1}{\sum_j e^{-\frac{1}{K \sqrt{n}}([1_K 1_K^T - \Phi] a)_j}} \left[ \sum_j [1_K 1_K^T - \Phi]_{ij} e^{-\frac{1}{K \sqrt{n}} ([1_K 1_K^T - \Phi] a)_j} \right].
\]
Note that this quantity is zero for $i=1$, and so $a_1$ stays at its initial value. It also does not appear in the evolution equations of the other variables, and so we drop it and only consider the non-trivial modes, viewing $a \in \mathbb{R}^{K-1}$ to be the vector containing only the non-trivial singular values.

The evolution equations are then:
\[
\frac{d}{dt}(a_i) = \frac{\sqrt{n}}{D} b_i, \qquad \textrm{where } b_i = \sum_{j=1}^{K-1} \Psi_{ij} e^{- \frac{1}{K \sqrt{n}} (\Psi a)_j}, \quad D= 1 + \sum_{j=1}^{K-1} e^{- \frac{1}{K \sqrt{n}} (\Psi a)_j}.
\]
Absorbing constants into $a$ and $t$ via: $a' = \frac{1}{K \sqrt{n}} a$ and $t'=\frac{1}{K}t$, then dropping primes since these re-scalings do not change the phenomenology of the dynamics, reduces us to
\[
\frac{d}{dt}(a_i) = \frac{1}{D} b_i, \qquad \textrm{where } b_i = \sum_{j=1}^{K-1} \Psi_{ij} e^{- (\Psi a)_j}, \quad D= 1 + \sum_{j=1}^{K-1} e^{- (\Psi a)_j}.
\]
If we define the variables $\hat{a}_i = a_i / \|a \|_1$, then these have evolution equation given by:
\[
\frac{d}{dt} \left( \hat{a}_i \right) = \frac{1}{D \| a \|_1} \left[ b_i - \hat{a}_i \sum_j b_j \right].
\]
Again defining the metric $M$ as
\[
M= \frac{1}{2} \sum_i \left[ \hat{a}_i - \frac{1}{K-1} \right]^2.
\]
This has derivative:
\[
\frac{dM}{dt} = \frac{1}{D \|a \|_1} \left[ \left(\sum_i \hat{a}_i b_i \right) - \left( \sum_i \hat{a}_i^2 \right) \left( \sum_j b_j \right) \right].
\]

Now note that, as in the proof of Theorem \ref{thm:NC_conv}, if we define $x_i = (Ta)_i$, we have:
\[
\sum_i \hat{a}_i b_i = \frac{1}{\| a\|_1} \sum_j x_j e^{-x_j},
\]
and as was shown in the proof, this satisfies:
\[
\frac{1}{\| a\|_1} \sum_j x_j e^{-x_j} \leq \frac{1}{(K-1) \|a\|_1} \left( \sum_i x_i \right)  \left( \sum_j e^{-x_j} \right),
\]
with equality only when $x \propto 1_{K-1}$. Hence we have that the derivative of $M$ satisfies:
\[
\frac{dM}{dt} \leq \frac{1}{D \|a \|_1} \left[ \frac{1}{(K-1) \|a \|_1} \left( \sum_i x_i \right)  \left( \sum_j e^{-x_j} \right) -  \left( \sum_i \hat{a}_i^2 \right) \left( \sum_j b_j \right)\right].
\]
Then use that $\sum_i b_i = K \sum_i e^{-x_i}$, and $\sum_i x_i = K \|a \|_1$, reducing this to
\[
\frac{dM}{dt} \leq \frac{K \sum_j e^{-x_j}}{D \|a \|_1} \left[ \frac{1}{(K-1)} -  \left( \sum_i \hat{a}_i^2 \right)\right].
\]
Then note by Cauchy-Schwarz we have $\sum_i \hat{a}_i^2 \geq \frac{1}{K-1}$, with equality only when $\hat{a} \propto 1_{K-1}$, consequently we arrive at 
\[
\frac{dM}{dt} \leq 0,
\]
with equality only when $x \propto 1_{K-1}$ and $\hat{a} \propto 1_{K-1}$. These both only occur when $\hat{a} = \frac{1}{K-1} 1_{K-1}$, as argued in the proof of Theorem \ref{thm:NC_conv}.

Hence the metric $M$ monotonically decreases to zero in this direct logit optimization case.

\subsection{Proof of Theorem \ref{thm:NC_basin}} \label{sec:NC_basin_proof}

Recall our metric $M$, given by:
\[
M= \frac{1}{2} \sum_{i=1}^{K-1} \left( \frac{a_i}{\| a\|_1} - \frac{1}{K-1} \right)^2\,.
\]
Throughout the proof we make use of the following properties of 
$\Psi = 1_{K-1}1_{K-1}^\top - X$: 
\begin{align*}
\Psi\,1_{K-1} &= K\,1_{K-1}, \\
\Psi^2 &= K\,I_{K-1} + K\,1_{K-1}1_{K-1}^\top, \\
|\Psi_{ij}| &\le 2, \qquad \forall\, i,j, \\
\|\psi_u\|_2 &= \sqrt{2K}, \qquad \forall \text{ columns } \psi_u \text{ of } \Psi.
\end{align*}
Additionally, unless otherwise stated, all indices run in $1,\ldots,k-1$ and are dropped for simplifying notation.

We choose to write our singular modes in the following form: 
\[
a_i = \mu + \mu \epsilon_i, \qquad \sum_i \epsilon_i=0, \qquad \mu = \frac{1}{K-1} \sum_i a_i\,.
\]
In this form $M$ is given by
\[
M = \frac{1}{2(K-1)^2} \sum_i \epsilon_i^2\,.
\]
Recall its derivative as given in Eq. \eqref{eq:metric_deriv}:
\[
\frac{dM}{dt} = \frac{1}{D} \sum_i \hat{a}_i^2[b_i - \bar{b}], \qquad \textrm{where } \bar{b} = \sum_j \hat{a}_j b_j\,.
\]
This we can write alternatively as
\begin{align*}
\frac{dM}{dt} &= \frac{1}{D} \left[\left(\sum_{i} \hat{a}_i^2 b_i \right) - \sum_{ij} \hat{a}_i^2 \hat{a}_j b_j \right]=\frac{1}{D \|a\|_1^3} \left[\left(\sum_{i} a_i^2 b_i \right) \left( \sum_j a_j \right) - \sum_{ij} a_i^2 a_j b_j \right]
\\&= \frac{1}{D \|a\|_1^3} \sum_{ij} a_i^2 a_j (b_i - b_j)\,.
\end{align*}

Now writing this in terms of $\epsilon$ and $\mu$ gives
$$\frac{dM}{dt} = \frac{1}{D \| a\|_1^3} \sum_{iju} (\mu + \mu \epsilon_i)^2 (\mu + \mu \epsilon_j) [ \Psi_{iu} - \Psi_{ju}]e^{-\mu(\Psi[ 1_{K-1} + \epsilon])_u }$$
$$ =  \frac{e^{-K \mu}}{D(K-1)^3} \sum_{iju} (1 + \epsilon_i)^2 (1+ \epsilon_j)[\Psi_{iu} - \Psi_{ju}] e^{- \mu (\Psi \epsilon)_u}\,,$$
where we used that $\mu = \frac{1}{K-1} \| a\|_1$, and $\Psi 1_{K-1} = K 1_{K-1}$.

Expand the exponential as:
$$\exp( - \mu (\Psi\epsilon)_u) = 1 -\mu (\Psi \epsilon)_u + \sum_{n=2}^{\infty} \frac{1}{n!} \left[ -\mu (\Psi \epsilon)_u \right]^n\,.$$

Using $\sum_u [\Psi_{iu} - \Psi_{ju}]=0$, the first term in $\frac{dM}{dt}$ gives
$$ \sum_{iju} (1 + \epsilon_i)^2 (1+ \epsilon_j)[\Psi_{iu} - \Psi_{ju}] = 0\,.$$ 

Further using that $\Psi^2 = KI_{K-1} + K 1_{K-1} 1_{K-1}^T$, the second term becomes
\begin{align*}
    - \mu \sum_{iju} (1 + \epsilon_i)^2 (1+ \epsilon_j)[\Psi_{iu} - \Psi_{ju}] (\Psi \epsilon)_u&= -\mu \sum_{ij} (1 + \epsilon_i)^2 (1+ \epsilon_j)[(\Psi^2 \epsilon)_i - (\Psi^2 \epsilon)_j]
    \\
    &=-K \mu \sum_{ij} (1 + \epsilon_i)^2 (1+ \epsilon_j)[ \epsilon_i - \epsilon_j]
    \\
    &=-K\mu \big[\sum_i(1+\eps_i)^2\eps_i\sum_j(1+\eps_j)-\sum_{i}(1+\eps_i)^2\sum_j(1+\eps_j)\eps_j\big]
    \\
    &=- K \mu \left[ (K-1) \sum_i \epsilon_i^2 + (K-1) \sum_i \epsilon_i^3 - \left( \sum_i \epsilon_i^2 \right)^2 \right]\,,
    \end{align*}
    where in the last line we used $\sum_i\eps_i=0$.

Again because $\sum_i\eps_i=0$ and $\sum_u\Psi_{iu}=K$, we can also simplify the remainder term by using
$$
\sum_{ij} (1 + \epsilon_i)^2(1 + \epsilon_j) (\Psi_{iu} - \Psi_{ju}) = \sum_i [(K-1) \Psi_{iu} \epsilon_i +(K-1) \Psi_{iu} \epsilon_i^2 - K \epsilon_i^2 - \epsilon_i^2 (\Psi \epsilon)_u]\,.
$$
Putting all the above together, our remaining expression for the full derivative is given by
\begin{align*}\frac{dM}{dt} &= -\frac{K \mu e^{-K \mu}}{D(K-1)^2} \sum_i \epsilon_i^2 - \frac{K \mu e^{-K \mu}}{D(K-1)^2} \sum_i \epsilon_i^3 + \frac{K \mu e^{-K \mu}}{D(K-1)^3} \left( \sum_i \epsilon_i^2 \right)^2 
\\&\qquad\quad+
\frac{ e^{-K \mu}}{D(K-1)^3} \sum_{iu} \left[(K-1) \Psi_{iu} \epsilon_i +(K-1) \Psi_{iu} \epsilon_i^2  \underbrace{-K \epsilon_i^2 - \epsilon_i^2 (\Psi \epsilon)_u}_{(*)} \right] \left[ \sum_{n=2}^{\infty} \frac{1}{n!} [- \mu (\Psi \epsilon)_u]^n \right]\,.
\end{align*}

We want to show that when $M$ is small enough, this quantity is negative. First let us assume that $| (\Psi \epsilon)_u| \leq K$, in which case we can drop the two terms highlighted as $(*)$, since they contribute a non-positive value, where we used that $e^x - 1 - x \geq 0$. Hence, also dropping a positive multiplicative constant, it remains to show

$$0 \geq -K \mu \sum_i \epsilon_i^2 - K \mu \sum_i \epsilon_i^3  + \frac{K \mu}{K-1} \left( \sum_i \epsilon_i^2 \right)^2 + \sum_{iu} (\Psi_{iu} \epsilon_i + \Psi_{iu} \epsilon_i^2)\left[ \sum_{n=2}^{\infty} \frac{1}{n!} [- \mu (\Psi \epsilon)_u]^n \right] \,.$$

Now we use the following properties to bound the remaining terms:
$$\sum_i \epsilon_i^2 = 2(K-1)^2 M,$$
$$|\sum_i \epsilon_i^3 |\leq \left( \sum_i \epsilon_i^2 \right)^{\frac{3}{2}}= 2^{\frac{3}{2}} (K-1)^3 M^\frac{3}{2},$$
$$\sum_i \Psi_{iu} \epsilon_i^2 \leq 2 \sum_i \epsilon_i^2 = 4(K-1)^2 M,$$
\begin{equation} \label{eq:phi_eps_bound}
    \sum_i \Psi_{iu} \epsilon_i = (\Psi \epsilon)_u \leq |(\Psi \epsilon)_u| \leq \sqrt{(\Psi \epsilon)\cdot(\Psi \epsilon)} = \sqrt{2K} (K-1) \sqrt{M}.
\end{equation}

Note the second property follows from monotonicity of $L_p$ norms, the third uses $\Psi_{iu}\in\{0,2\}$ for all $i,u,$ and, the last one $\|\psi_u\|= \sqrt{2 K}$ for all columns/rows $\psi_u$ of $\Psi$. This produces the following sufficient condition for a negative derivative:
\begin{align*}
0 &\geq -2K \mu (K-1)^2 M + 2^{\frac{3}{2}} K \mu (K-1)^3 M^\frac{3}{2} + 4K \mu (K-1)^3 M^2 
\\
&\qquad\qquad+ \sum_u \left[ \sqrt{2K} (K-1) \sqrt{M} + 4 (K-1)^2 M \right] \underbrace{\left[ \sum_{n=2}^{\infty} \frac{1}{n!} [- \mu (\Psi \epsilon)_i]^n \right]}_{R_u}\,.
\end{align*}
We can bound the quantity $R_u$ using $e^x - 1 - x \leq \frac{1}{2} x^2 e^{|x|}$, giving 
$$R_u \leq \frac{1}{2} \mu^2 (\Psi \epsilon)_u^2 \exp(| \mu (\Psi \epsilon)_u |) \leq K \mu^2 (K-1)^2 M \exp \left( \mu \sqrt{2K} (K-1) \sqrt{M} \right)\,.$$
where in the second step we used the bound for $(\Psi \epsilon)_u$ from Eq. \eqref{eq:phi_eps_bound}. Hence, it is sufficient to show
\begin{align*}0 &\geq -2K \mu (K-1)^2 M + 2^{\frac{3}{2}} K \mu (K-1)^3 M^\frac{3}{2} + 4K \mu (K-1)^3 M^2 
\\ &\quad\qquad + 
\left[ \sqrt{2K} (K-1)^2 \sqrt{M} + 4 (K-1)^3 M \right] K \mu^2 (K-1)^2 M \exp \left( \mu \sqrt{2K} (K-1) \sqrt{M} \right)\,,
\end{align*}
where we also picked up an extra factor of $(K-1)$ due to the summation over $u$.

Now, dropping a factor of $2K(K-1)^2 \mu M$, this reduces to
\begin{align*}
0 \geq -1 + 2^{\frac{1}{2}} (K-1) M^\frac{1}{2} + 2 (K-1) M + 
\left[ \sqrt{\frac{K}{2}} (K-1)^2 \sqrt{M} + 2 (K-1)^3 M \right] \mu \exp \left( \mu \sqrt{2K} (K-1) \sqrt{M} \right)\,.
\end{align*}

Letting $\tilde{M} = 2KM$ and $\tilde{\mu} = (K-1) \mu$, this further reduces to
$$0 \geq -1 + \frac{K-1}{\sqrt{K}} \tilde{M}^{\frac{1}{2}} + \frac{K-1}{K} \tilde{M} + \left[ \frac{1}{2} (K-1) \tilde{M}^{\frac{1}{2}} + \frac{(K-1)^2}{K} \tilde{M} \right] \tilde{\mu} e^{\tilde{\mu} \tilde{M}^\frac{1}{2}}\,,$$
and finally using that $K-1 \leq K$ it is sufficient to show
$$1 \geq \sqrt{K} \tilde{M}^{\frac{1}{2}} + \tilde{M} + \left[ \frac{1}{2} \tilde{M}^{\frac{1}{2}} + \tilde{M} \right]K \tilde{\mu} e^{\tilde{\mu} \tilde{M}^{\frac{1}{2}}}\,.$$

Now assume $\tilde{M}^{\frac{1}{2}} \leq \min \left( \frac{1}{2\sqrt{K}}, \frac{1}{3K \tilde{\mu}} \right)$, then we have:

$$\sqrt{K} \tilde{M}^{\frac{1}{2}} + \tilde{M} + \left[ \frac{1}{2} \tilde{M}^{\frac{1}{2}} + \tilde{M} \right]K \tilde{\mu} e^{\tilde{\mu} \tilde{M}^{\frac{1}{2}}} \leq \frac{1}{2} + \frac{1}{4K} + \left[ \frac{1}{2} + \frac{1}{2\sqrt{K}} \right] \frac{1}{3} e^{\frac{1}{3K}} \leq\frac{1}{2} + \frac{1}{8} + \left[ \frac{1}{2} + \frac{1}{2\sqrt{2}} \right] \frac{1}{3} e^{\frac{1}{6}} \approx 0.961\,,$$
where we used that $K \geq 2$. This quantity is less than one, and hence $\frac{dM}{dt}$ is negative subject to this constraint on $\tilde{M}$. Stated in terms of $M$ and $\| a\|_1$, this constraint is
$$M \leq \min \left( \frac{1}{8K^2} , \frac{1}{18K^3 \|a\|_1^2} \right)\,.$$

Recall that we also used $|(\Psi \epsilon)_u| \leq K$, previously, but this is a weaker bound than the above (by Eq. \eqref{eq:phi_eps_bound}) and so holds subject to our condition. Hence we have shown that 

$$\frac{dM}{dt} \leq 0, \quad \textrm{when } M \leq \min \left( \frac{1}{8K^2} , \frac{1}{18K^3 \|a\|_1^2} \right)\,.$$

\subsection{Proof of Theorem \ref{thm:initialization}}

We now consider randomly initialized matrices for the UFM model of Eq. \eqref{eq:CE_UFM}. Assume that $W,H$ are initialized as $W(0)=\epsilon d^{-\frac{1}{2}} A_1$, $H(0)= \epsilon d^{-\frac{1}{2}} A_0$, so that the logit matrix is given by $Z(0) = \epsilon^2 d^{-1} A_1 A_0$.

We assume that for a large value $t^* \in \mathbb{R}$ the value of $\epsilon$ is small enough that for the time window $t \in [0,t^*]$, $W(t)=O(\epsilon)$, $H(t) = O( \epsilon)$, and so $Z(t) = O(\epsilon^2)$, meaning we remain in this lower order approximate regime on that time window. Note by $O(\cdot)$, we mean $\exists C(t^*)>0$ such that $\| Z(t,\epsilon) \|_F \leq C(t^*) \epsilon^2$ for $t \in [0,t^*]$. We will capture the relationship between $t^*$ and $\epsilon$ later in the proof.

Hence for this time window, we have that the softmax is given by:
\[
P(t)=\frac{1}{K} 1_K 1_{Kn}^T + P'(t, \epsilon),
\]
where $P'(t, \epsilon) = O(\epsilon^2)$. From the gradient flow equations for this model, described in Eq. \eqref{eq:grad_flow}, we have for $t \in [0,t^*]$
\[
\frac{dW}{dt} = (S \otimes 1_n^T)H^T - P' H^T, \qquad \frac{d H^T}{dt} = (S \otimes 1_n^T)^T W - P' W.
\]
We can reduce this to a second order ODE
\[
\frac{d^2W}{dt^2} = nSW + \underbrace{\frac{d}{dt}(-P' H^T) - (S \otimes 1_n^T)P'W}_{O(\epsilon^3)},
\]
where we have the boundary conditions $W(0) = \epsilon d^{-\frac{1}{2}} A_1$, $\frac{dW}{dt}(0) = \epsilon d^{-\frac{1}{2}} (S \otimes 1_n^T)A_0^T + O(\epsilon^3)$. We also note that the terms that feature $P'$ are $O( \epsilon^3)$ on the considered time interval, so should be viewed as perturbative.

This has solution given by
\[
W(t)=\epsilon d^{-\frac{1}{2}} \left[ \frac{1}{K} 1_K 1_{K}^T A_1 + \cosh(t \sqrt{n})SA_1 + \frac{1}{\sqrt{n}} \sinh(t \sqrt{n})(S \otimes 1_n^T)A_0^T \right] + O(\epsilon^3 t^2),
\]
and corresponding solution for $H$ given by
\[
H^T(t) = \epsilon d^{-\frac{1}{2}} \left[(I_{Kn} - \frac{1}{n} S \otimes 1_n 1_n^T)A_0^T + \frac{1}{\sqrt{n}} \sinh(t \sqrt{n}) (S \otimes 1_n) A_1 + \frac{1}{n} \cosh (t \sqrt{n}) (S \otimes 1_n 1_n^T)A_0^T \right] + O(\epsilon^3 t^2).
\]
We now want to compute $Z(t^*)=W(t^*) H(t^*)$. In this expansion we recover terms of a variety of different orders in $t$ and $\epsilon$. We now define 
$$t^* = \frac{1}{2\sqrt{n}} \log \left( \frac{1}{\epsilon} \right)$$

This diverges in the $\epsilon \to 0$ limit, but slowly enough that the logit matrix norm goes to zero, meaning our perturbative terms remain small. By considering the terms in $W(t)$ and $H(t)$, it becomes clear that the leading order term in $Z(t^*)=W(t^*)H(t^*)$ comes from the product of the terms in $W,H$ that are exponential in $t^*$, giving
\[
Z(t^*) = \frac{1}{4d} \epsilon^2 e^{2t^* \sqrt{n}} \left[ SA_1 + \frac{1}{\sqrt{n}} (S \otimes 1_n^T) A_0^T \right]  \left[ \frac{1}{\sqrt{n}} (S \otimes 1_n)A_1 + \frac{1}{n} (S \otimes 1_n 1_n^T) A_0^T \right]^T + \dots\, ,
\]
where by $\dots$ we mean terms that vanish faster than leading order. Continuing 
\[
Z(t^*) = \frac{1}{4d} \epsilon^2 e^{2t^* \sqrt{n}} S\left[ A_1 + \frac{1}{\sqrt{n}} (I_K \otimes 1_n^T) A_0^T \right]  \left[ A_1 + \frac{1}{\sqrt{n}} (I_K \otimes 1_n^T) A_0^T \right]^T (S \otimes \frac{1}{\sqrt{n}} 1_n^T) + \dots \,.
\]
Now define the matrix $X = \frac{1}{\sqrt{2}} [A_1 + (I_K \otimes \frac{1}{\sqrt{n}} 1_n^T)A_0^T]$, this gives
\[
Z(t^*) = \frac{1}{2d} \epsilon^2 e^{2t^* \sqrt{n}} S X X^T (S \otimes \frac{1}{\sqrt{n}} 1_n^T) + \dots \, .
\]
Now suppose that the entries of $A_1 \in \mathbb{R}^{K \times d}$ and $A_0^T \in \mathbb{R}^{Kn \times d}$ are initialized with i.i.d. samples from a standard Gaussian. Then the entries of the matrix $\frac{1}{\sqrt{n}}(I_K \otimes 1_n^T) A_0^T \in \mathbb{R}^{K \times d}$ are also i.i.d. samples of a standard Gaussian, and hence the matrix $X \in \mathbb{R}^{K \times d}$ also has entries i.i.d. sampled from a standard Gaussian.

Considering the entries of the matrix $\frac{1}{d} X X^T$ in the large $d$ limit, we find:
\[
i=j: \ \left[ \frac{1}{d} (XX^T) \right]_{ii} = \frac{1}{d} \sum_u X_{iu} X_{iu} \rightarrow \mathbb{E} [N_1^2] = 1 ,
\]
\[
i \neq j: \ \left[ \frac{1}{d} (XX^T) \right]_{ij} = \frac{1}{d} \sum_u X_{iu} X_{ju} \rightarrow \mathbb{E} [N_1 N_2] = \mathbb{E}[N_1] \mathbb{E}[N_2] = 0 .
\]
Where $N_1,N_2 \sim \mathcal{N}(0,1)$.

Hence in the large $d$ limit $\frac{1}{d} X X^T \to I_K$, and we find that 
\[
\frac{1}{d} S X X^T (S \otimes \frac{1}{\sqrt{n}} 1_n^T) \approx \frac{1}{\sqrt{n}} (S \otimes 1_n^T),
\]
and so $Z(t^*)$ is well approximated as
\[
Z(t^*) = \frac{1}{2 \sqrt{n}} \epsilon^2 e^{2t^* \sqrt{n}} S \otimes 1_n^T + \dots \, ,
\]
where the higher order terms decay faster in the described limits, and so this gives the asymptotic leading term.

\section{Additional Experiments}\label{sec:details exp}

\subsection{Metrics}
In the numerical simulations discussed in Sec. \ref{sec:experiments}, we track the following metrics.
\begin{itemize}
    \item[{(a)}] \emph{Cross-Entropy Loss:} Plots the  CE loss in Eq. \eqref{eq:CE_UFM}.

    \item[{(b)}] \emph{Alignment Metrics:} Tracks the Frobenius norm distance between the nuclear-norm-normalized model components ($WW^T, H^T H, Z=WH$) and the ground-truth Simplex ETF $S$. For example for the logit matrix $Z=WH$ it plots 
    $\left\| \frac{Z}{\|Z\|_*} - \frac{S}{\|S\|_*} \right\|_F$. For Hadamard Init., where $Z$ and $S$ share singular spaces, this is identical to the (square root) of the metric $M$ in Eq. \eqref{eq:metric}.

    \item[{(c)}] \emph{KL Divergence:} Plots the Lyapunov function from Theorem \ref{thm:NC_conv}, measuring the KL divergence of the normalized logit singular values $\hat{{a}}$ from the uniform distribution $\frac{1}{K-1}{1}_{K-1}$.

    \item[{(d)}] \emph{Singular Values:} Plots the evolution of the raw singular values $\sigma_i(Z)$ of the logit matrix $Z=WH$. For the \texttt{SVD} and \texttt{ODE} runs, these are equivalent to the paper's logit singular values $a_i$.

    \item[{(e)}] \emph{Normalized Singular Values:} Plots the $L_1$-normalized singular values $\hat{\sigma}_i(Z) = \sigma_i(Z) / \sum_j \sigma_j(Z)$. For the \texttt{SVD} and \texttt{ODE} runs, these are equivalent to the paper's $\hat{a}_i = a_i / \|a\|_1$.

    \item[{(f)}] \emph{Residual Energy:} Measures the fraction of the logit matrix $Z$ that lies outside the  signal subspace of the Hadamard Init.. Specifically, for $U=V=\Phi(1:K,2:K)$, it measures
    ${\| Z - {U} ({U}^T Z {V}) {V}^T \|_F}\big/{\|Z\|_F}$
\end{itemize}

For efficiency,  we track and plot these metrics in $100$ logarithmically spaced time instants.

\subsection{Larger $K$}
\label{app:k8_sim}

\begin{figure}[h]
    \centering
    \includegraphics[width=0.9\textwidth]{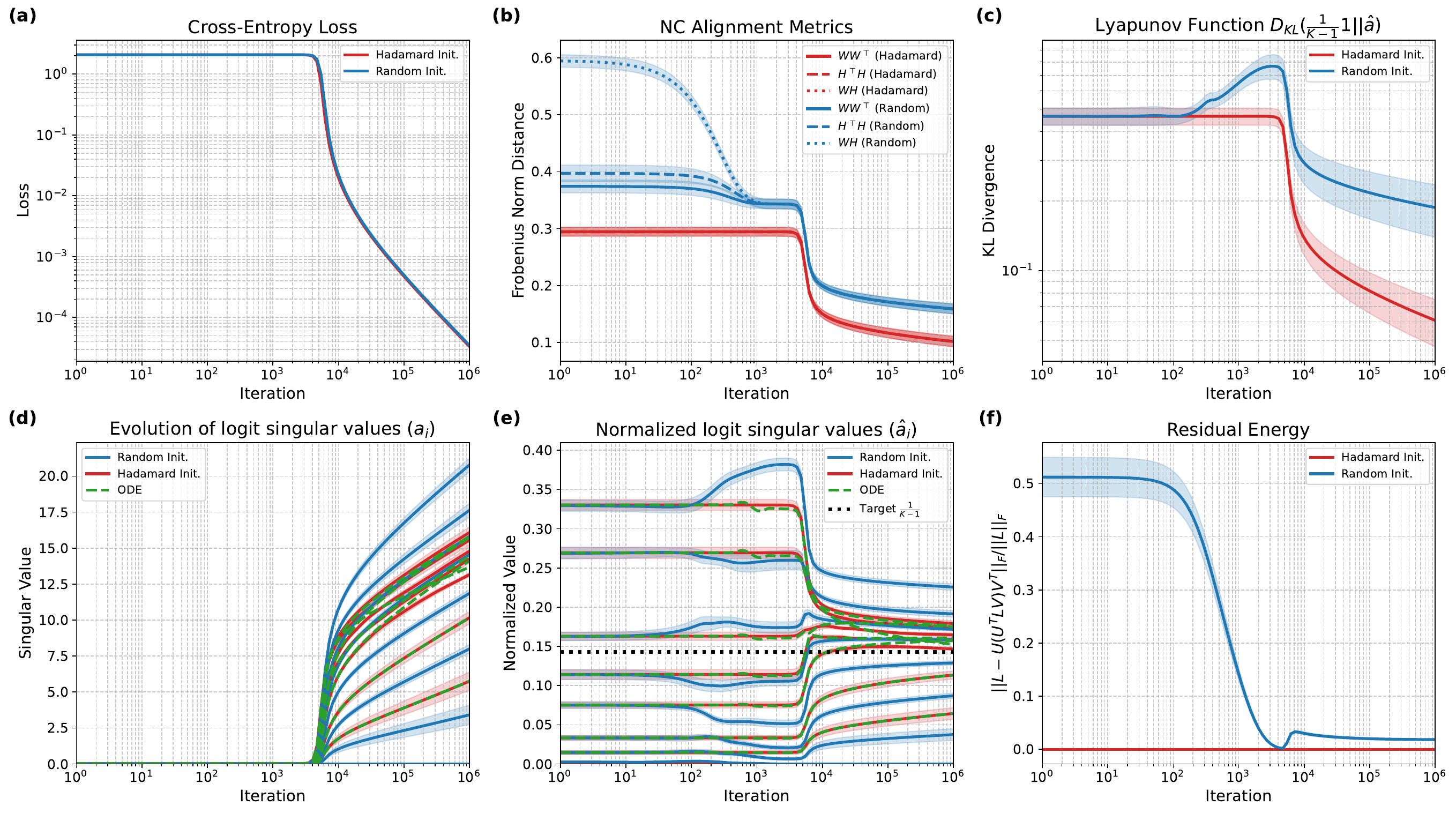}
    \caption{
        {Comparison of Hadamard Init., Random Init., and theoretical ODE for $K=8$ classes.} 
    }
    \label{fig:K8_sim}
\end{figure}

\begin{figure}[h]
    \centering
    \includegraphics[width=0.9\textwidth]{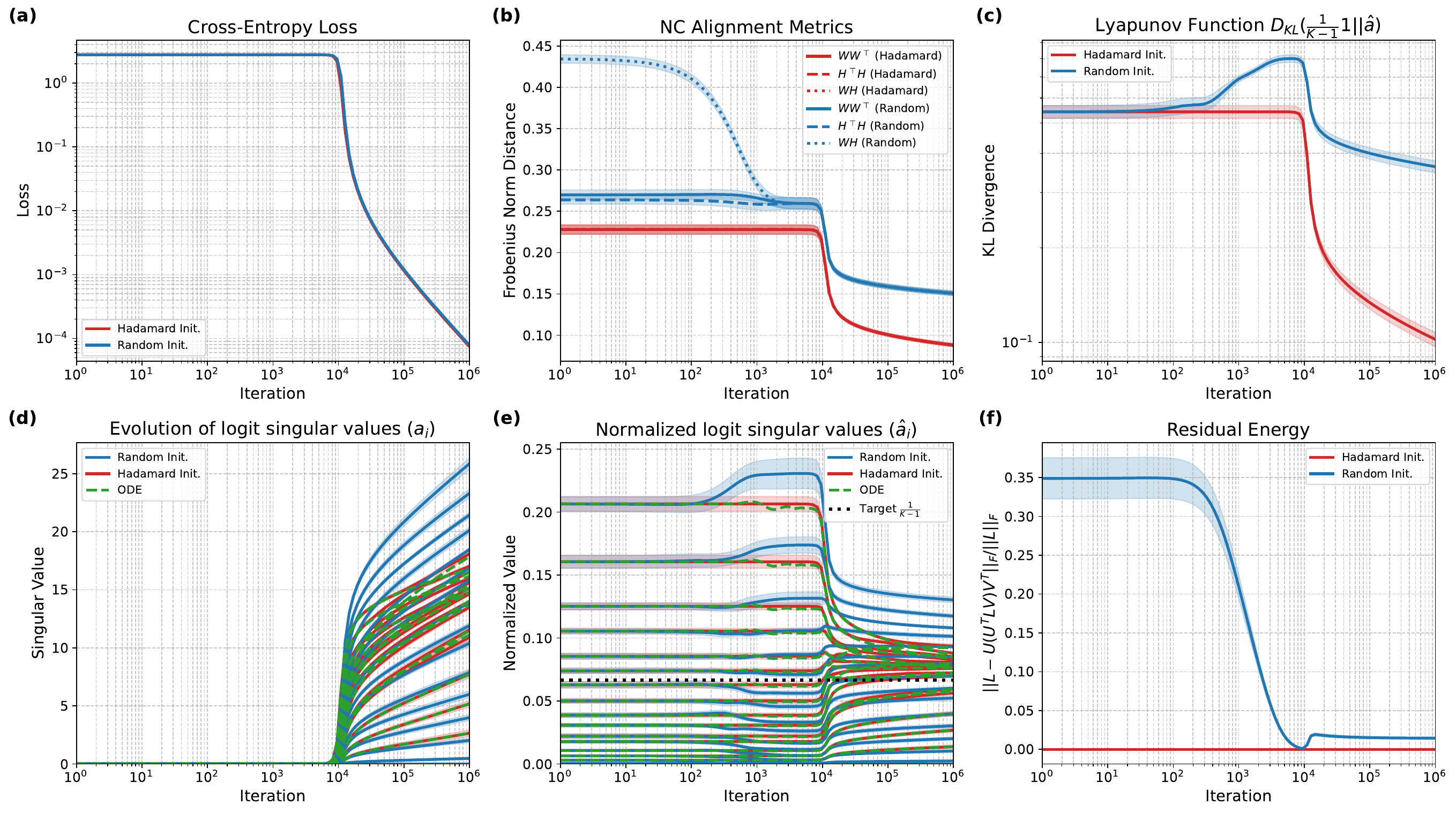}
    \caption{
        {Comparison of Hadamard Init., Random Init., and theoretical ODE for $K=16$ classes.}
    }
    \label{fig:K16_sim}
\end{figure}

We present the simulation results for $K=8$ classes in Fig.~\ref{fig:K8_sim}. 
Many of the core findings from the $K=4$ case remain:
\begin{itemize}[leftmargin=*]
    \item \emph{Fixed Vectors:} The Residual Energy (plot f) for the Hadamard Init. (red) is zero for all time, confirming the singular vectors are fixed.
    \item \emph{ODE Validation:} The ODE (green dashed) perfectly tracks the Hadamard Init. dynamics (red) in plots (d) and (e).
\item \emph{Two-Phase Dynamics:} The Random Init. (blue) exhibits a distinct "burn-in" phase (up to $t \approx 5 \times 10^3$) where its singular vectors align with the ETF subspace (plot f) while the loss remains flat (plot a). During this alignment, the normalized singular values of the Hadamard Init. remain flat. It is only after this burn-in phase that the normalized singular values for both Random and Hadamard Initializations begin converging toward the NC equilibrium.
\end{itemize}

However, note two differences in the $K=8$ case. First, the alignment of the normalized singular values between the two initializations, even after the burn-in phase, is weaker than in the $K=4$ case. Second, it is evident from plots (c) and (e) that the convergence to NC is significantly slower for $K=8$, even for the Hadamard Initialization.

The observations above appear consistent for $K=16$ as shown in Figure \ref{fig:K16_sim}.

\begin{figure}[h]
    \centering
    \includegraphics[width=0.9\textwidth]{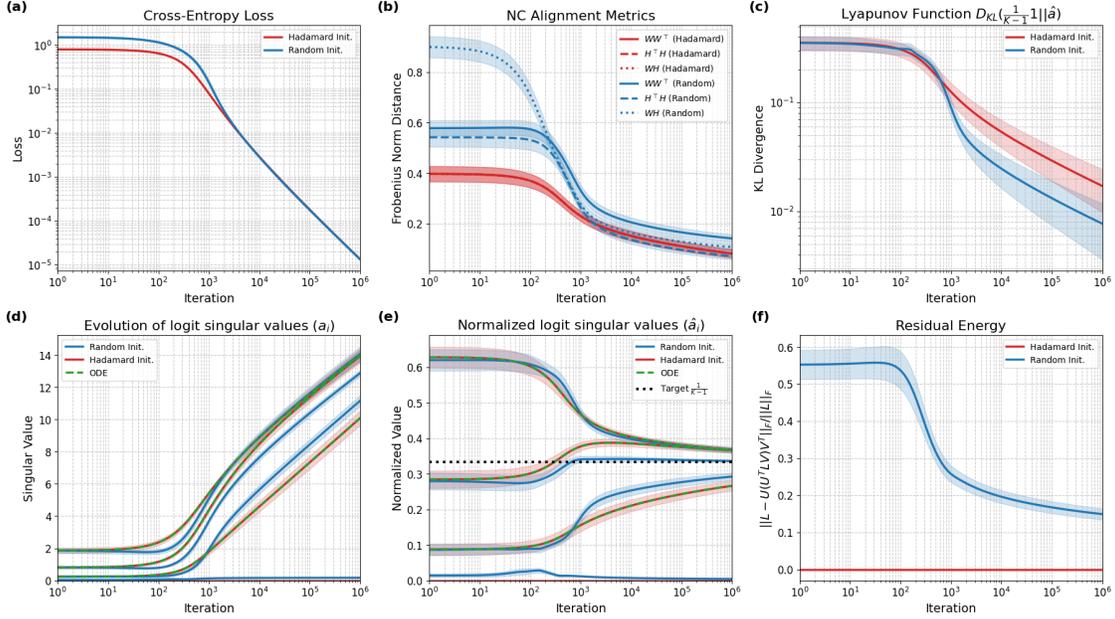}
    \caption{
        {Same as Figure \ref{fig:V=4_svd_vs_rand} for $K=4$ but now with standard initialization scale (rather than infinitesimal).}
    }
    \label{fig:K4_sim_delta=0}
\end{figure}
\begin{figure}[h]
    \centering
    \includegraphics[width=0.9\textwidth]{Plots/UFM/sim_V8_R1_N10_iter1000000_lr0.01_delta0_initSameSpec_logs100.pdf}
    \caption{
        Same as Figure \ref{fig:K8_sim} for $K=8$ but now with standard initialization scale (rather than infinitesimal).
    }
    \label{fig:K8_sim_delta=0}
\end{figure}

\subsection{The effect of initialization scale}
Thus far we used small initializations: recall from Sec.~\ref{sec:experiments} that $W$ and $H$ are initialized with i.i.d.\ Gaussian entries of variance $1/K$ and then scaled by $\epsilon=e^{-\delta}$ with $\delta=6$. For MSE, following Saxe et al.~\cite{Saxe2019}, this vanishing initialization, yields dynamics similar to spectral initialization; our previous CE experiments were the analogous setting. For completeness, we also report results under a standard scale, i.e., $\epsilon=1$ ($\delta=0$). Figures~\ref{fig:K4_sim_delta=0} and \ref{fig:K8_sim_delta=0} show $K=4$ and $K=8$, respectively. The two-stage dynamics seen with infinitesimal initialization persist, but with quantitative changes: e.g., the initial “burn-in” shortens to $t\!\sim\!5\times10^{2}$. Notably, while the alignment of logit singular values to their NC equilibrium (measured by NC metrics and KL divergence) remains comparable to the infinitesimal case, the residual energy is markedly higher: subspace alignment from Random Initialization  to the simplex ETF subspace proceeds much more slowly.

\subsection{The effect of overparameterization}

In the experiments thus far, we have set the UFM's hidden dimension to $d=K$. Since the simplex ETF subspace has dimension $K-1$, this is essentially the minimal choice ensuring that the simplex ETF structure is attainable. For completeness, we examine the overparameterized regime with $d>K$.

\begin{figure}[h]
    \centering
    \includegraphics[width=0.9\textwidth]{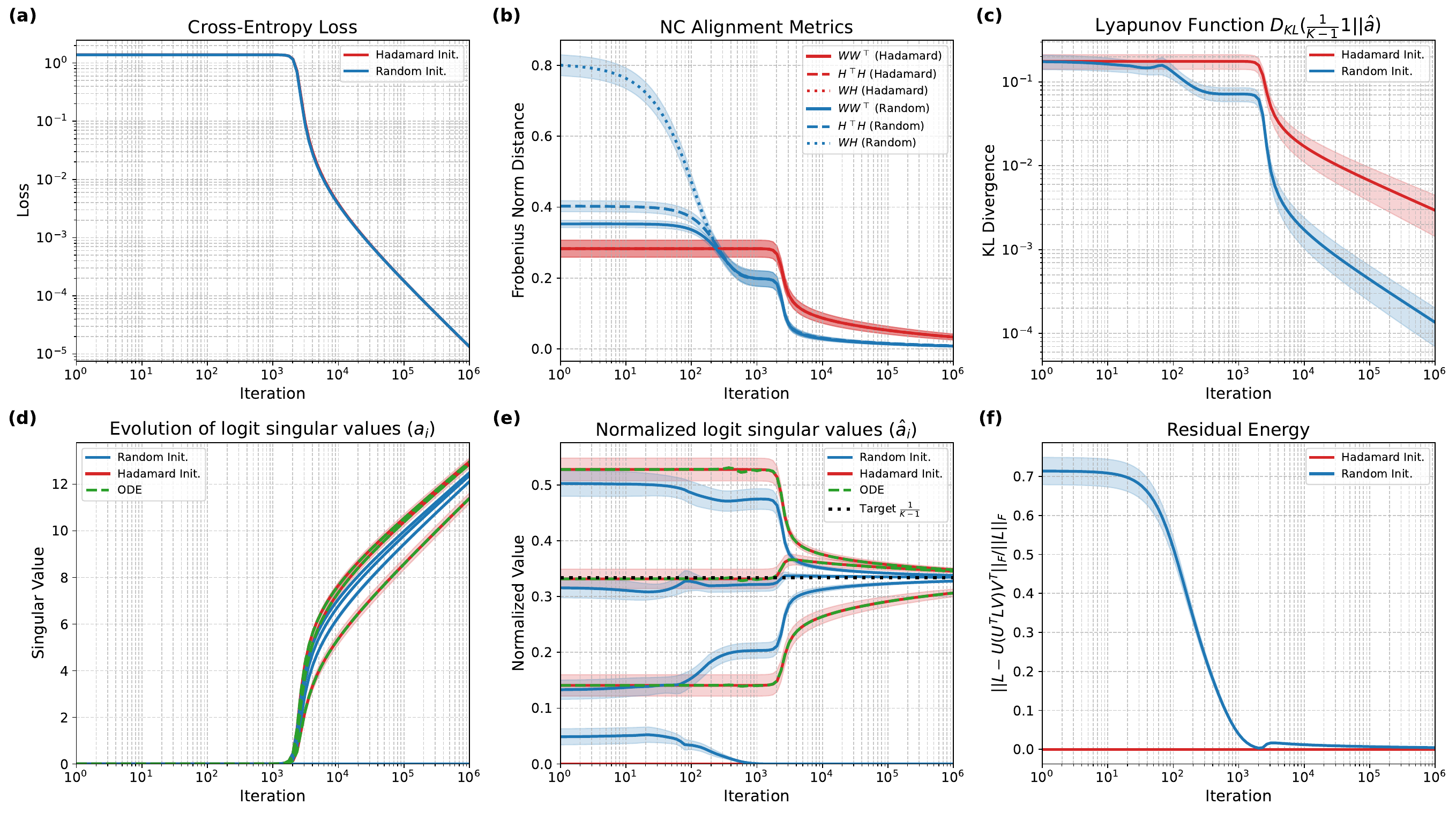}
    \caption{
        Same as Figure \ref{fig:V=4_svd_vs_rand} for $K=4$ but now with hidden dimension $d=5\cdot K$.
    }
    \label{fig:K4_sim_dx5}
\end{figure}

\begin{figure}[h]
    \centering
    \includegraphics[width=0.9\textwidth]{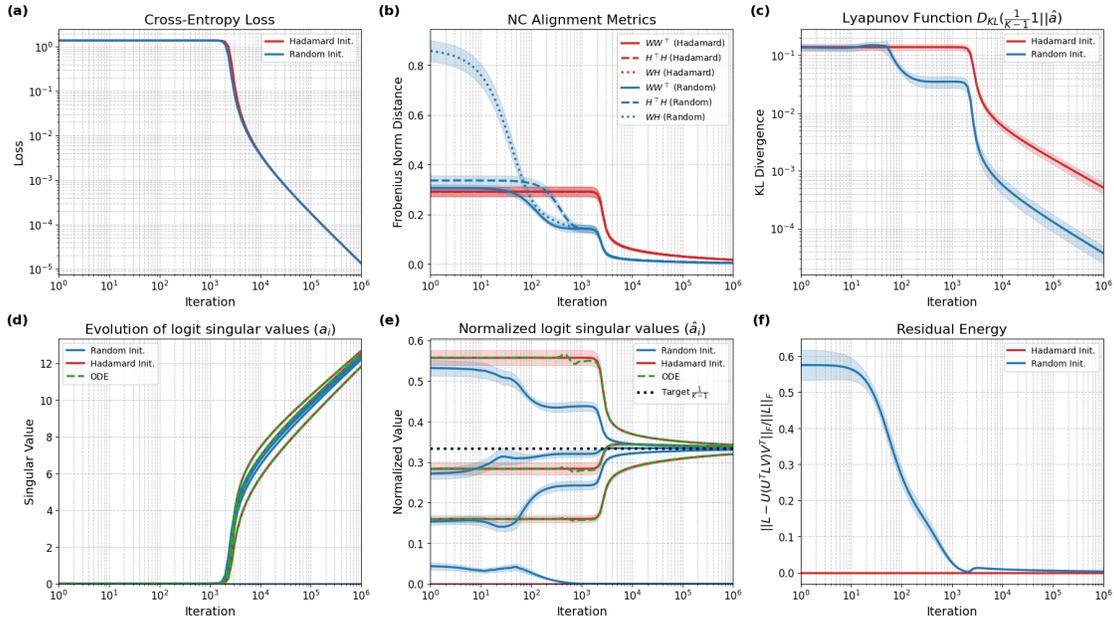}
    \caption{
        Same as Figure \ref{fig:V=4_svd_vs_rand} for $K=8$ but now with hidden dimension $d=10\cdot K$.
    }
    \label{fig:K4_sim_dx10}
\end{figure}

For Hadamard Initialization, our theory predicts that $d$ should not affect the dynamics. Figures \ref{fig:K4_sim_dx5} and \ref{fig:K4_sim_dx10}, showing overparameterization factors $d/K=5$ and $10$ respectively, confirm this prediction. 

However, overparameterization benefits Random Initialization: while the loss plateau persists, the burn-in phase (where parameter magnitudes remain small) now exhibits rapid convergence of normalized singular values toward equilibrium. Interestingly, this occurs despite the residual energy decaying at roughly the same rate as in the $d=K$ case, though it remains closer to zero during the second phase, avoiding the slight increase observed without overparameterization. 

This behavior aligns with the predictions of Theorem \ref{thm:initialization}: Under infinitesimal random initialization, zero-th order UFM dynamics already drive the solution to simplex ETF geometry.


\subsection{Normalized Gradient Flow}\label{sec:normalized exp ode}

We have seen that gradient flow converges to the NC solution (with all normalized singular values $\hat{a}_i$ equal), but convergence can be very slow. The counterexamples constructed for metric $M$ in Sec. \ref{sec:non_monotone}—specifically Figure \ref{fig:non_monotone_conv}—illustrate this. For the initialization shown, metric $M$ increases, but Theorem \ref{thm:NC_conv} guarantees it must eventually decrease to 0 as the flow converges to equilibrium. However, this occurs extremely slowly. To numerically verify this behavior, we implement an accelerated version of the reduced flow in Eq. \eqref{eq:grad_flow}, the \emph{normalized} gradient flow:
\begin{align}
    \frac{da}{dt} = \frac{F(a)}{\|F(a)\|_2}\qquad \text{where } F(a)=a\odot \nabla_a E(a) = \frac{a\odot b}{D}. \label{eq:ngd reduced}
\end{align}
For convex models, $L_2$-normalized flow accelerates optimization without modifying convergence to the same max-margin solution as standard flow \cite{nacson2019convergence}. We leave proving this property in our non-convex setting to future work.

Unfortunately, direct implementation of the updates in Eq. \eqref{eq:ngd reduced} encounters numerical issues: The gradients of the energy functional $E(a)$ involve terms that can overflow when margins become large during training. This is particularly worrisome for normalized flow, which accelerates margin growth. To implement gradient flow stably, we rewrite gradient terms by factoring out dominant exponential terms and subtracting the minimum margin, an idea inspired by \cite[App.~I]{lyu2019gradient}.

First note that given the gradient components $F_i(a) = a_i b_i / D$, the denominator $D$ cancels entirely:
\begin{align*}
\frac{da_i}{dt} = \frac{F_i(a)}{\|F(a)\|_2} = \frac{a_i b_i / D}{\sqrt{\sum_k (a_k b_k / D)^2}} =\frac{a_i b_i}{\| a \odot b \|_2}\,.
\end{align*}

We now apply a stability trick to the vector $b$ itself. Let $m_j = (\Psi a)_j$ be the margins and $m_{\min} = \min_j m_j$. We define a numerically stable rescaled vector $\hat{b}$ as:
\begin{align}
\hat{b}_i := b_i \cdot e^{m_{\min}} = \left( \sum_{j=1}^{K-1} \Psi_{ij} e^{-m_j} \right) e^{m_{\min}} = \sum_{j=1}^{K-1} \Psi_{ij} e^{-(m_j - m_{\min})}\,.
\end{align}
Substituting $b_i = \hat{b}_i e^{-m_{\min}}$ back into the normalized dynamics, the large exponential terms cancel:
\begin{align*}
\frac{da_i}{dt} = \frac{a_i (\hat{b}_i e^{-m_{\min}})}{\| a \odot (\hat{b} e^{-m_{\min}}) \|_2} =  \frac{a_i \hat{b}_i}{\| a \odot \hat{b} \|_2}\,.
\end{align*}
This gives the final, stable formulation for the normalized dynamics:
\begin{align}
\frac{da_i}{dt} = \frac{a_i \sum_{j=1}^{K-1} \Psi_{ij} e^{-(m_j - m_{\min})}}{\sqrt{\sum_k \left( a_k \sum_{j=1}^{K-1} \Psi_{kj} e^{-(m_j - m_{\min})} \right)^2}}
\end{align}
In this form, all exponents are non-positive, with at least one exponent in the sums being exactly zero. This is numerically stable for large values of $a$.

\begin{figure}[t!]
  \centering
  \includegraphics[width=1\textwidth]{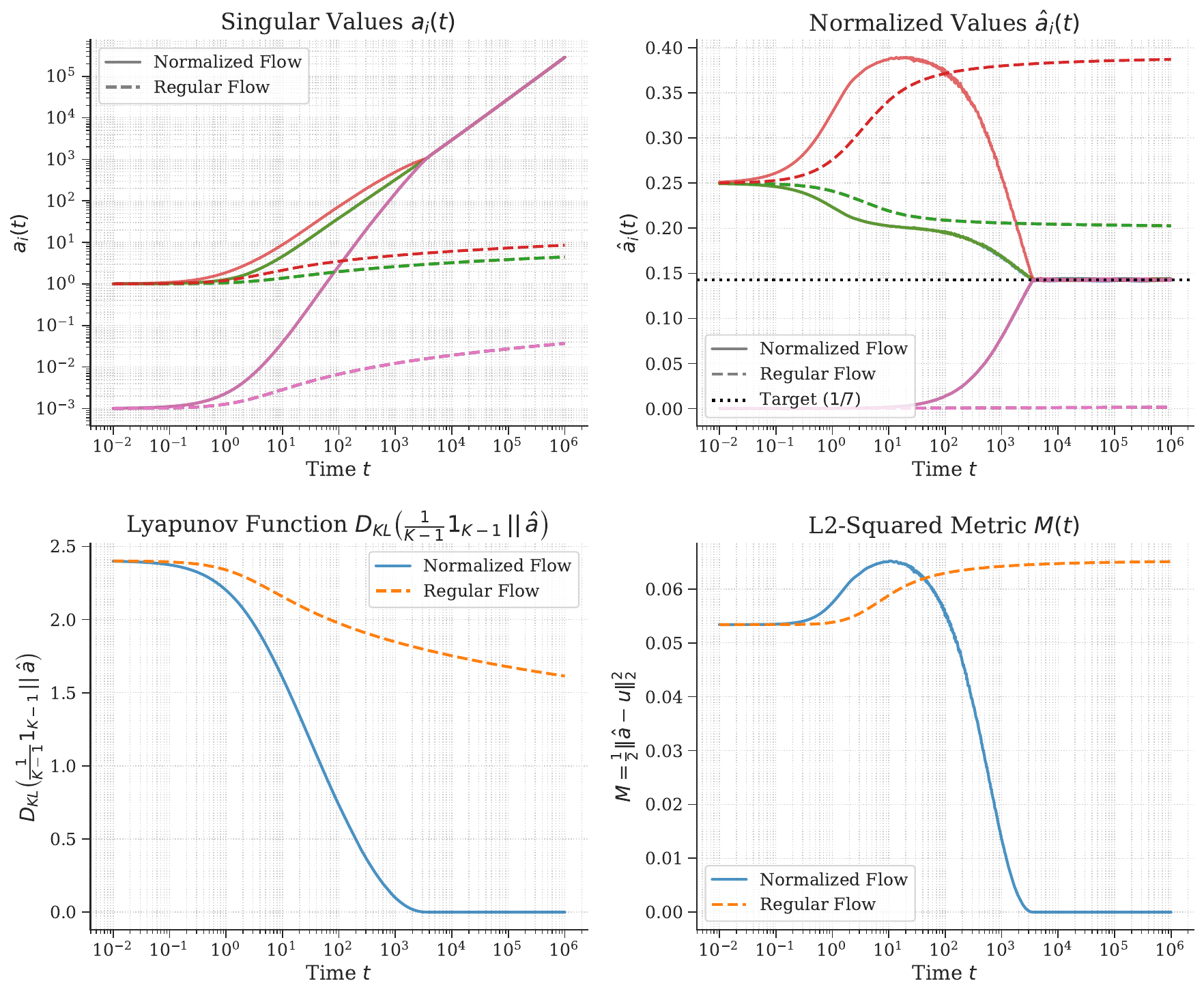}
  \caption{
    \textbf{Normalized Flow vs. Regular ODE Flow ($K=8$).}
    A direct comparison of the normalized flow (solid lines) and the regular stable ODE flow (dashed lines) for the highly non-uniform initialization $a(0)=[1,1,1,1,0.001,0.001,0.001]$.
  }
  \label{fig:flow_comparison}
\end{figure}

Figure \ref{fig:flow_comparison} compares the normalized flow (solid lines) with regular gradient flow (dashed lines) for the non-monotonic initialization from Figure \ref{fig:non_monotone_conv}. The acceleration is dramatic: while regular flow requires time scales beyond $t=10^6$ to approach the NC equilibrium $\hat{a}_i = 1/7$ (top right), normalized flow achieves this by path length $\tau \approx 10^3$. Both the Lyapunov function $D_{\mathrm{KL}}$ (bottom left) and metric $M$ (bottom right) converge orders of magnitude faster under normalized flow. 

The non-monotonic behavior of metric $M$ observed in Figure \ref{fig:non_monotone_conv} remains visible: normalized flow (solid blue) exhibits a similar but compressed trajectory where $M$ first increases before eventually decreasing. Notably, the singular values (top left) become extremely large when the decrease in $M$ begins. Without the stable implementation above, we would have encountered numerical overflow at this stage. This confirms that normalized flow preserves the qualitative dynamics while making convergence tractable on practical time scales.

\end{document}